\documentclass[letterpaper, 10 pt, conference]{ieeeconf}  

\IEEEoverridecommandlockouts                              

\usepackage{makecell}
\usepackage{amsmath} 
\usepackage{amssymb}  

\usepackage{amsfonts}
\usepackage{amssymb}
\usepackage{amsmath}
\usepackage{graphics} 
\usepackage{epsfig} 
\usepackage{mathptmx} 
\usepackage{times} 
\usepackage{subfigure}

\usepackage{mathrsfs}
\usepackage{indentfirst}
\usepackage{multirow}

\usepackage{cite}

\usepackage{algorithmic}
\usepackage{graphicx}
\usepackage{textcomp}
\usepackage{xcolor}
\usepackage{bbding}
\usepackage{geometry}
\usepackage[colorlinks=true,linkcolor=blue,citecolor=green,urlcolor=blue,]{hyperref}

\title{\LARGE \bf
	M2DGR: A Multi-sensor and Multi-scenario SLAM Dataset \\
	for Ground Robots}

\author{Jie Yin,  Ang Li, Tao Li, Wenxian Yu, and Danping Zou$^{*}$,
	\thanks{All authors are with Shanghai Key Laboratory of Navigation and Location Based Services,
		Shanghai Jiao Tong University. This work was supported by NSFC(62073214).}%
	\thanks{$^*$ Corresponding Author: Danping Zou ({\tt\small dpzou@sjtu.edu.cn})}
}

\begin{document}

	\maketitle
	\thispagestyle{empty}
	\pagestyle{empty}


	\begin{figure*}[t]
		\begin{center}
			\footnotesize
			\begin{tabular}{cc}
				\includegraphics[scale=0.19]{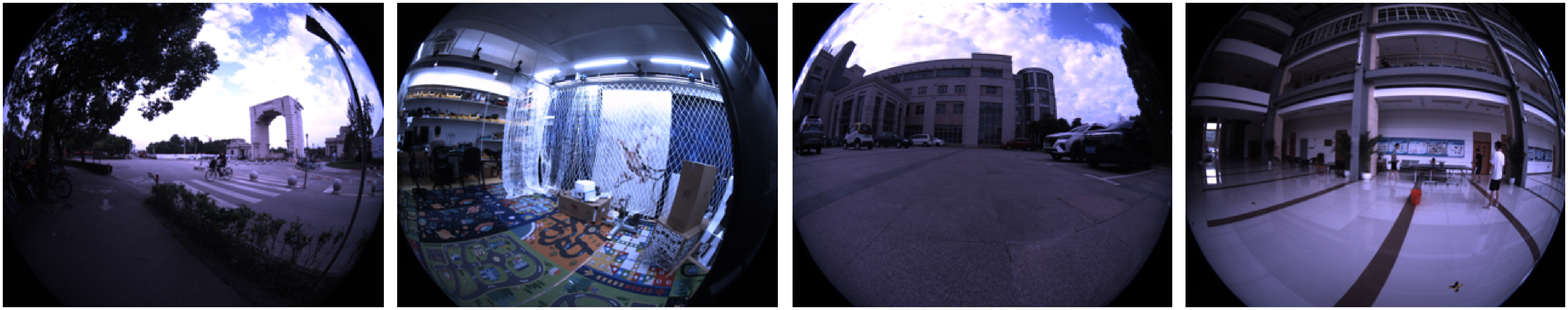} &
				\includegraphics[scale=0.19]{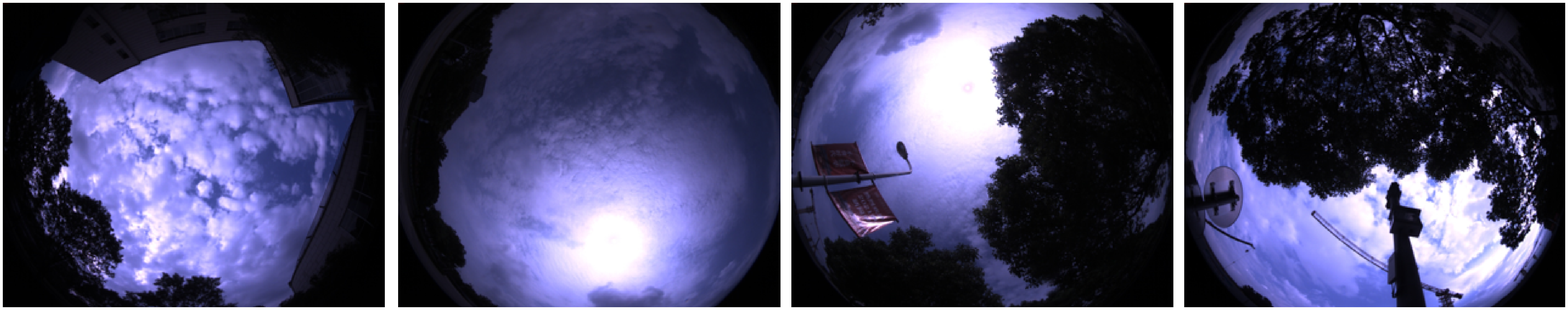} 
				\\
				(a) Surround-view fish-eye cameras & (b) Sky-pointing camera\\
				
				\includegraphics[scale=0.19]{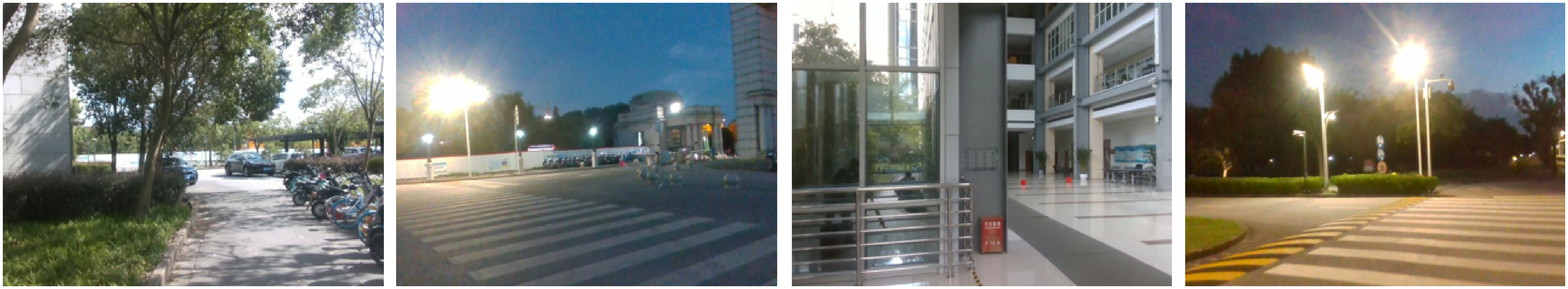}&
				\includegraphics[scale=0.19]{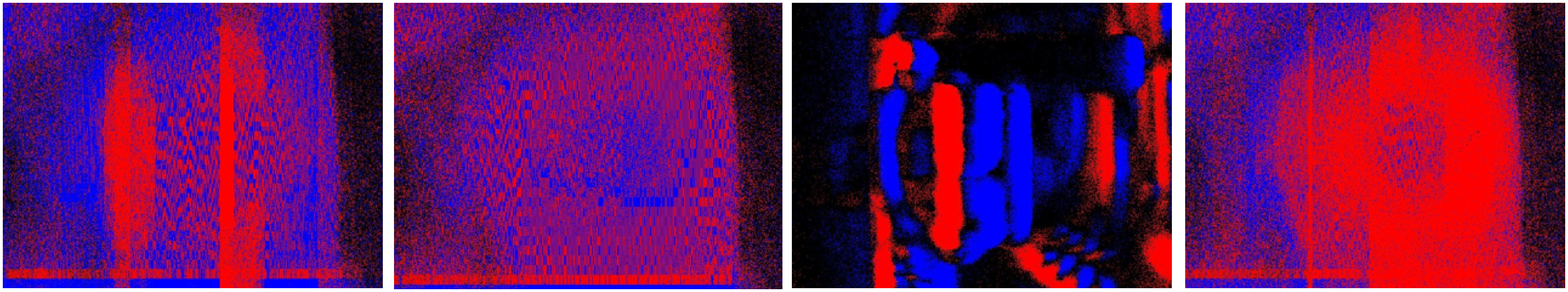}
				\\
				(c) Forward-looking pinhole camera& (d) Event camera\\
				
				\includegraphics[scale=0.19]{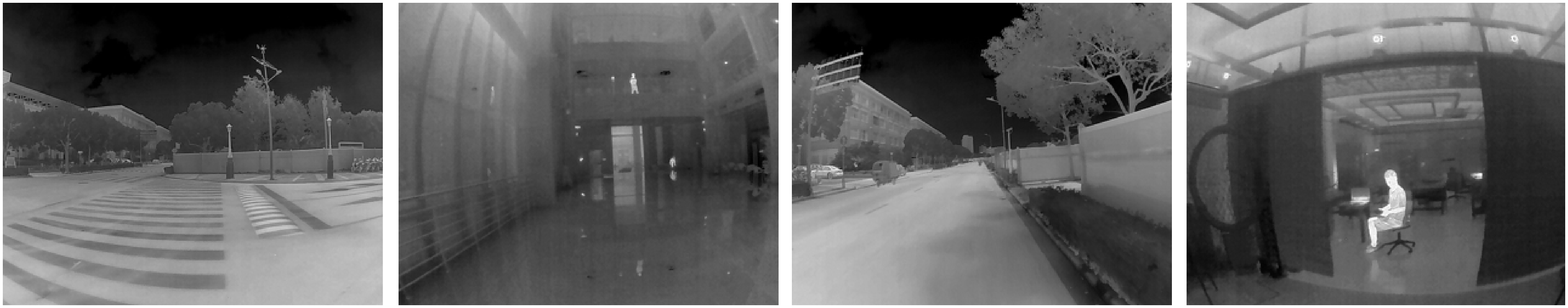}&
				\includegraphics[scale=0.19]{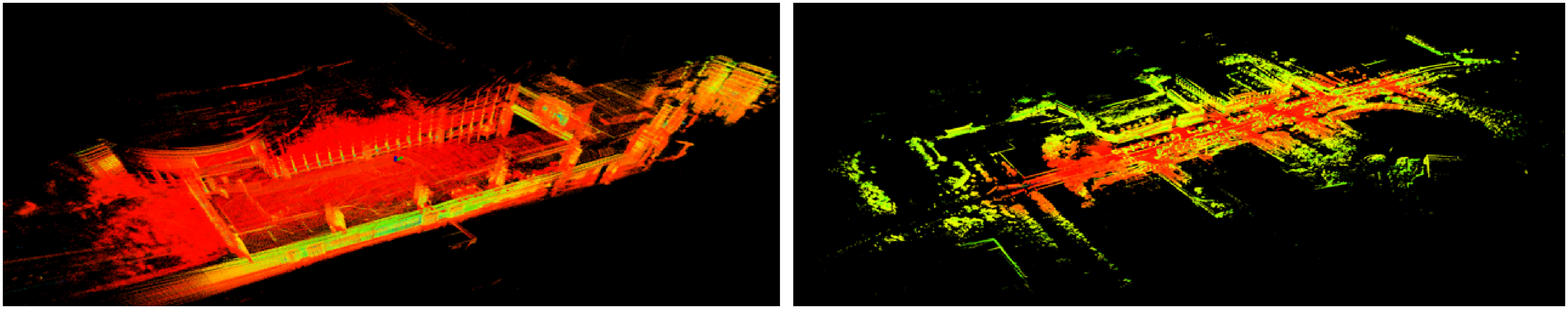}
				\\
				(e)Thermal-infrared camera& (f) LiDAR \\
			\end{tabular}
		\end{center}
		\caption{Our dataset for ground robots was captured by a rich suite of sensors within various scenarios. Some sensory data are visualized.}
		\label{ate rmse fig}
	\end{figure*}

	\begin{abstract}
		
		We introduce M2DGR: a novel large-scale dataset collected by a ground robot with a full sensor-suite including six fish-eye and one sky-pointing RGB cameras, an infrared camera, an event camera, a Visual-Inertial Sensor (VI-sensor), an inertial measurement unit (IMU), a LiDAR, a consumer-grade Global Navigation Satellite System (GNSS) receiver and a GNSS-IMU navigation system with real-time kinematic (RTK) signals. All those sensors were well-calibrated and synchronized, and their data were recorded simultaneously. The ground truth trajectories were obtained by the motion capture device, a laser 3D tracker, and an RTK receiver.  The dataset comprises 36 sequences (about 1TB) captured in diverse scenarios including both indoor and outdoor environments. We evaluate state-of-the-art SLAM algorithms on M2DGR. Results show that existing solutions perform poorly in some scenarios. For the benefit of the research community, we make the dataset and tools public. The webpage of our project is  \href{https://github.com/SJTU-ViSYS/M2DGR}{https://github.com/SJTU-ViSYS/M2DGR}.
	\end{abstract}
	
	\section{INTRODUCTION}
	Intelligent ground robots have been emerging in a wide range of applications such as logistics, security, warehouses, cleaning, and food delivery \cite{appleton2012industrial
}. In those applications, the robots need to work reliably in indoor or a mixture of indoor and outdoor scenes. SLAM (Simultaneous Localization and Mapping) is the critical enabling technology that allows those robots to navigate in those complex scenes, which can construct a map of the environment while simultaneously tracking the location of the robot within the map. Though SLAM research has made a substantial progress in the past decades \cite{aulinas2008slam},\cite{cadena2016past}, existing solutions frequently perform poorly in practice.  For example, visual SLAM may fail at texture-less or dark backgrounds, while LiDAR SLAM could have trouble with long corridors or open areas. SLAM may also become invalid when the robot takes some unusual actions, for instance, a robot moves into a lift and goes out to the new floor. Those failure cases motivate us to construct a dataset that includes more practical scenarios to facilitate SLAM research.

High-quality datasets can speed up breakthroughs and enable a fair comparison between different algorithms. However, most existing SLAM datasets are designed for autonomous driving or aerial robots as pointed in \cite{liu2021datasets}. Those datasets are not the best fit for developing and evaluating algorithms of ground robots. Autonomous cars travel fast on streets and roads, while the ground robots move at a much lower speed in remarkably different surroundings, including both indoor and outdoor scenes. The aerial robots fly freely in 3D space and are also quite different from ground robots. Although there are a few datasets targeting ground robots \cite{sturm2012benchmark},\cite{shi2020we}, they include only a few specific sensors or particular scenes. For logistics robots, catering robots, and service robots, challenging scenarios are frequently faced, like going into a lift or complete darkness, or going from outdoors to indoors. These situations may easily make existing SLAM methods fail, while they are seldom included in existing SLAM benchmark tests.

In this paper, we introduce a new dataset for SLAM research of ground robots, which includes both indoor and outdoor environments and contains a rich suite of sensors. The dataset contains trajectories in highly diverse scenes such as halls, lifts, corridors, and roads. Based on this dataset, we evaluate the state-of-the-art SLAM algorithms, including both LiDAR SLAM and visual SLAM. The results show that existing SLAM systems perform poorly in at least one situation, indicating further efforts are required to improve the SLAM performance. We summarize major contributions as follows:

\begin{itemize}
	\item We collected large-scale sequences for ground robots with a rich sensor suite, which includes six surround-view fish-eye cameras, a sky-pointing fish-eye camera, a VI-sensor, an event camera, an infrared camera, a 32-beam LiDAR, an IMU, and two GNSS receivers. To our knowledge, this is the first SLAM dataset focusing on ground robot navigation with such rich sensory information.
	\item We recorded trajectories in challenging situations like entering lifts and complete darkness which are commonly faced in practical applications, whereas they are not present in previous datasets.
	\item We launched a comprehensive benchmark where we evaluated existing state-of-the-art SLAM algorithms of various designs and analyzed their characteristics and defects.
\end{itemize}
	
	\section{RELATED WORK}
	\subsection{SLAM with different sensors}
	
	Generally speaking, SLAM can be categorized into vision-based and laser-based ones. Vision-based SLAM, or visual SLAM, can be divided into monocular, binocular, and multi-camera settings according to the number of cameras \cite{cadena2016past}. Though monocular visual SLAM  \cite{qin2018vins} \cite{wang2018cubemapslam} is the most studied topic, it suffers from scale uncertainty and scale drift. Binocular visual SLAM\cite{campos2021orb} \cite{qin2019general} takes advantage of known baseline distance to calculate metric depth by triangulation, and multi-camera SLAM \cite{urban2016multicol} \cite{ang2021robust} yields a broader field of view by using more cameras to enhance the robustness of SLAM in dynamic environments such as streets. However, it is difficult for pure vision-based SLAM to handle scenes with few textures or low illumination conditions. By contrast, LiDAR SLAM is often regarded as a more reliable choice in such challenges. Nonetheless, LiDAR SLAM also has trouble with long corridors, highly dynamic movements, and foggy scenes. 
	
	Recently, multi-sensor fusion has been successfully applied to existing SLAM systems to improve both the accuracy and robustness in practice.  For example, ORB-SLAM3 \cite{campos2021orb} and VINS-Mono \cite{qin2018vins} integrate vision and IMU. LVI-SAM \cite{shan2021lvi}, and R2live \cite{lin2021r2live} tightly integrate vision, LiDAR, and IMU. LIO-SAM\cite{shan2020lio} loosely integrates LiDAR, IMU, and GNSS. GVINS \cite{cao2021gvins} tightly couples image information, IMU, and GNSS. 
	
	To further improve the SLAM performance in practice, a recent trend is to explore different sensor configurations. We list some of them as follows.
	
	\paragraph{Multiple cameras} Multiple cameras have a broader field of view than monocular or stereo cameras, which can improve the robustness in dynamic scenes \cite{zhang2016benefit}. CoSLAM \cite{zou2012coslam} uses independently moving multiple cameras and gains the capability of co-localization and robustness in dynamic movements. If multiple cameras are mounted on the same platform, not only a wider field of view can be achieved, but also the scale ambiguity can be resolved by the baselines between cameras \cite{ang2021robust}. For example, Multicol-SLAM \cite{urban2016multicol} applies three fixed fish-eye cameras on a helmet, ROVO \cite{seok2019rovo} uses four fish-eye cameras to achieve full coverage of 360° field of views. 
	Panoramic SLAM \cite{ji2020panoramic} is another open-source omnidirectional SLAM system, which claims to achieve centimeter-level accuracy even in highly dynamic movements.

	\paragraph{Thermal-infrared cameras}Thermal-infrared cameras have gained increasing attention for their perceptual capability beyond the visible spectrum and their robustness regarding environmental changes. In environments with low visibility such as fog, smoke, and darkness, visual SLAM with ordinary RGB image information becomes ineffective. By contrast, thermal-infrared images can effectively improve visibility in these scenarios. For example, J. Delaune \cite{delaune2019thermal} proposed a SLAM algorithm using thermal-infrared images, enabling autonomous flight of UAVs (Unmanned Aerial Vehicle) at night. 
	
	\paragraph{Event cameras} Event cameras measure changes in the brightness of pixels which are with low delay, low power consumption, and high dynamic measurement range. Therefore, they have unique advantages in quick motions. Henri et al. proposed an event-based Visual Odometry algorithm \cite{rebecq2016evo}. A. R. Vidal et al. proposed to tightly integrate events, images, and inertial information \cite{vidal2018ultimate} to achieve better performance.
	
	\paragraph{GNSS} GNSS is a valuable localization source that can achieve high-precision positioning outdoors. Coupling GNSS raw measurements into SLAM systems has been proven effective in advancing the localization performance in urban canyons, as shown in recent work\cite{li2020p}. Besides, a sky-pointing camera can help monitor satellite availability and further improve localization accuracy\cite{marais2014toward}.

	\begin{table*}
		\caption{Comparison of SLAM datasets}
		\label{dataset comparison}
		\centering
		\begin{tabular}{p{2.2cm}p{1.75cm}p{1.95cm}p{1.6cm}p{1.35cm}p{1cm}p{0.95cm}p{1.05cm}p{1.12cm}p{1cm}p{1cm}}
			\hline
			\makecell[c]{Dataset} & \makecell[c]{Environment} & \makecell[c]{Platform}  & \makecell[c]{Duration}  & \makecell[c]{RGB Cam.} & \makecell[c]{LiDAR} & \makecell[c]{IMU}  & \makecell[c]{Infrared} & \makecell[c]{GNSS} & \makecell[c]{Event}\\
			\hline
			\makecell[c]{KITTI \cite{geiger2013vision}}  & \makecell[c]{Urban} & \makecell[c]{Car}\ & \makecell[c]{Short-term$^{\mathrm{a}}$} & \makecell[c]{2} & \makecell[c]{\Checkmark} & \makecell[c]{\Checkmark}  & \makecell[c]{ } & \makecell[c]{ } & \makecell[c]{ }\\
			\makecell[c]{UrbanLoco \cite{wen2020urbanloco}}  & \makecell[c]{Urban}   & \makecell[c]{Car}    & \makecell[c]{Long-term}   & \makecell[c]{6} &  \makecell[c]{\Checkmark} & \makecell[c]{\Checkmark} &  \makecell[c]{ } & \makecell[c]{\Checkmark}  & \makecell[c]{ } \\
			\makecell[c]{Brno Urban  \cite{ligocki2020brno} }& \makecell[c]{Urban}   & \makecell[c]{Car}   & \makecell[c]{Long-term}   & \makecell[c]{4}  & \makecell[c]{\Checkmark} & \makecell[c]{\Checkmark} &  \makecell[c]{\Checkmark} &\makecell[c]{\Checkmark}  & \makecell[c]{ }\\
			\makecell[c]{Kaist D/N \cite{choi2018kaist}}  & \makecell[c]{Urban}   & \makecell[c]{Car}    & \makecell[c]{Long-term}     & \makecell[c]{2}  & \makecell[c]{\Checkmark} & \makecell[c]{\Checkmark} &  \makecell[c]{\Checkmark} & \makecell[c]{ } & \makecell[c]{ }\\
			
			\makecell[c]{Pit30M \cite{martinez2020pit30m}}  & \makecell[c]{Urban}   & \makecell[c]{Car} & \makecell[c]{Long-term}  & \makecell[c]{1}    & \makecell[c]{\Checkmark} & \makecell[c]{\Checkmark} &  \makecell[c]{ } & \makecell[c]{ } & \makecell[c]{ }\\ 
			
			\makecell[c]{USVinland \cite{cheng2021we}}  & \makecell[c]{Inland}   & \makecell[c]{USV}   & \makecell[c]{Short-term}    & \makecell[c]{2}  & \makecell[c]{\Checkmark} & \makecell[c]{ } &  \makecell[c]{ } & \makecell[c]{ } & \makecell[c]{ }\\
			
			\makecell[c]{EUROC \cite{burri2016euroc}}  &\makecell[c]{Indoors} & \makecell[c]{UAV}   & \makecell[c]{Short-term}    & \makecell[c]{2}   & \makecell[c]{ }  & \makecell[c]{\Checkmark} &  \makecell[c]{ } & \makecell[c]{ } & \makecell[c]{ }\\
			\makecell[c]{UZH-FPV \cite{delmerico2019we}}  & \makecell[c]{In/Outdoors} &\makecell[c]{UAV} & \makecell[c]{Short-term}   & \makecell[c]{2}    & \makecell[c]{ } & \makecell[c]{\Checkmark} &  \makecell[c]{ } & \makecell[c]{ } & \makecell[c]{\Checkmark}\\
			
			\makecell[c]{TUM VI \cite{schubert2018tum}}  &\makecell[c]{In/Outdoors} & \makecell[c]{Hand-held}   & \makecell[c]{Short-term}   & \makecell[c]{2}   & \makecell[c]{ }  & \makecell[c]{\Checkmark} &  \makecell[c]{ } & \makecell[c]{ } & \makecell[c]{ } \\
			
			\makecell[c]{LaFiDa \cite{urban2017lafida}}  & \makecell[c]{In/Outdoors} & \makecell[c]{Helmet}  & \makecell[c]{Short-term}    & \makecell[c]{3}   & \makecell[c]{ } & \makecell[c]{\Checkmark} &  \makecell[c]{ } & \makecell[c]{ }  & \makecell[c]{ } \\

			\makecell[c]{NCLT \cite{carlevaris2016university}}  & \makecell[c]{In/Outdoors}   & \makecell[c]{Ground Robot} & \makecell[c]{Long-term}  & \makecell[c]{6}    & \makecell[c]{\Checkmark} & \makecell[c]{\Checkmark} &  \makecell[c]{ } & \makecell[c]{ } & \makecell[c]{ }\\ 
			\makecell[c]{OpenLORIS \cite{shi2020we}}  &\makecell[c]{Indoors} & \makecell[c]{Ground Robot}   & \makecell[c]{Short-term}   & \makecell[c]{2}   & \makecell[c]{\Checkmark}  & \makecell[c]{\Checkmark} &  \makecell[c]{ } & \makecell[c]{ } & \makecell[c]{ } \\
			
			\hline
			\makecell[c]{\textbf{Our dataset}} & \makecell[c]{In/Outdoors/\\Transition/Lift}  & \makecell[c]{Ground Robot}  & \makecell[c]{Long-term}  & \makecell[c]{8} & \makecell[c]{\Checkmark} & \makecell[c]{\Checkmark} & \makecell[c]{\Checkmark} & \makecell[c]{\Checkmark} & \makecell[c]{\Checkmark}\\
			\hline     
			\multicolumn{9}{l}{$^{\mathrm{a}}$We identify a dataset as long-term if it has sequences longer than 20 minutes.}
		\end{tabular}
	\end{table*}

	\subsection{Existing benchmark datasets}
	
	\paragraph{Datasets for ground robots}
	Most existing SLAM datasets focus on autonomous driving \cite{wen2020urbanloco},\cite{geiger2013vision},\cite{ligocki2020brno} or UAVs \cite{burri2016euroc}. A few datasets are targeted at ground robots. OpenLORIS\cite{shi2020we} was collected by a wheeled robot in indoor environments, which was designed for visual SLAM, where LiDAR SLAM was used to generate the ground truth. As we will see later, LiDAR SLAM may have even more significant errors than visual SLAM in some situations, making the ground truth unreliable. TUM RGBD \cite{sturm2012benchmark} partly used a robot as the acquisition platform, but it only contains RGB and depth cameras. Similar datasets include UTIAS MultiRobot \cite{leung2011utias} and PanoraMIS \cite{benseddik2020panoramis}, which are not applicable to LiDAR SLAM as well. The aforementioned datasets have limitations such as lack of rich sensory sources, outdated data, and insufficient challenges.

	\begin{figure*}
		\centering
		\includegraphics[scale=0.27]{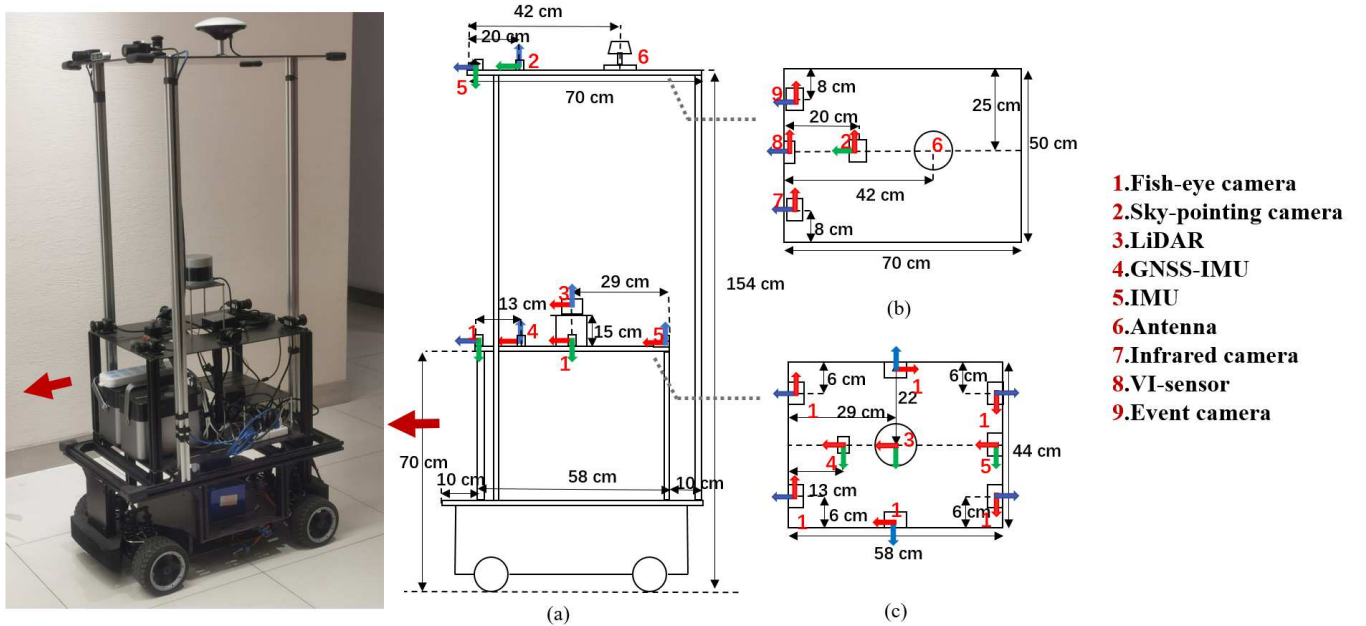}
		\caption{Our ground robot for data collection. (a) Side view of the robot. (b) Sensors on the top layer. (c) Sensors on the middle layer.}
		\label{car}
	\end{figure*}
	
	\paragraph{Datasets with multiple cameras}
	Lafida \cite{urban2017lafida} dataset contains three fish-eye cameras on a helmet, but the maximum recording time of its sequences is too short for long-term evaluation (usually longer than 20 minutes). The NCLT \cite{carlevaris2016university} dataset used an omnidirectional camera on a ground robot to capture images on campus at a frame rate of 5Hz. Such a low frame rate could cause a small overlap between adjacent frames, which will be problematic to run visual SLAM algorithms. Nuscenes \cite{caesar2020nuscenes}, Waymo \cite{sun2020scalability}, and A2D2 \cite{geyer2020a2d2} datasets collected image data from multiple cameras with a 360° field of view in urban areas, but they did not provide ground truth of trajectories for evaluation of SLAM performance.
	
	\paragraph{Datasets with thermal-infrared camera}
	Brno Urban dataset recorded the infrared camera information on a car \cite{ligocki2020brno}. The KAIST D/N \cite{choi2018kaist} dataset collected data from a stereo camera and a thermal infrared camera on a car, as well as a 32-beam LiDAR. However, as far as we know, no public SLAM dataset contains the data of infrared cameras in indoor scenes, which could be useful for the research and development of indoor navigation algorithms in the night or the smoke.
	\begin{figure*}
		\begin{center}
			\begin{tabular}{cccc}
				\includegraphics[scale=0.08]{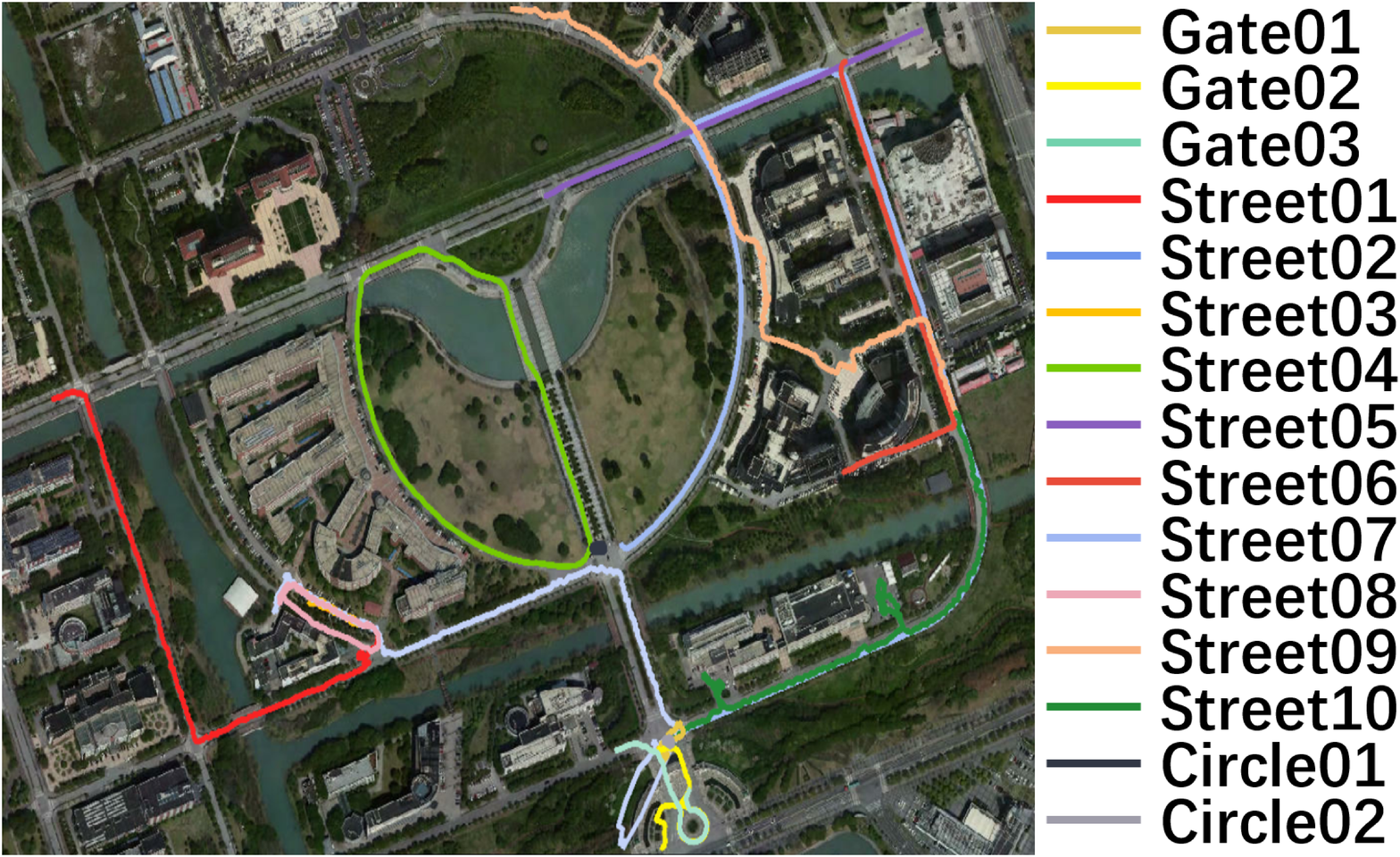} &
				\includegraphics[scale=0.044]{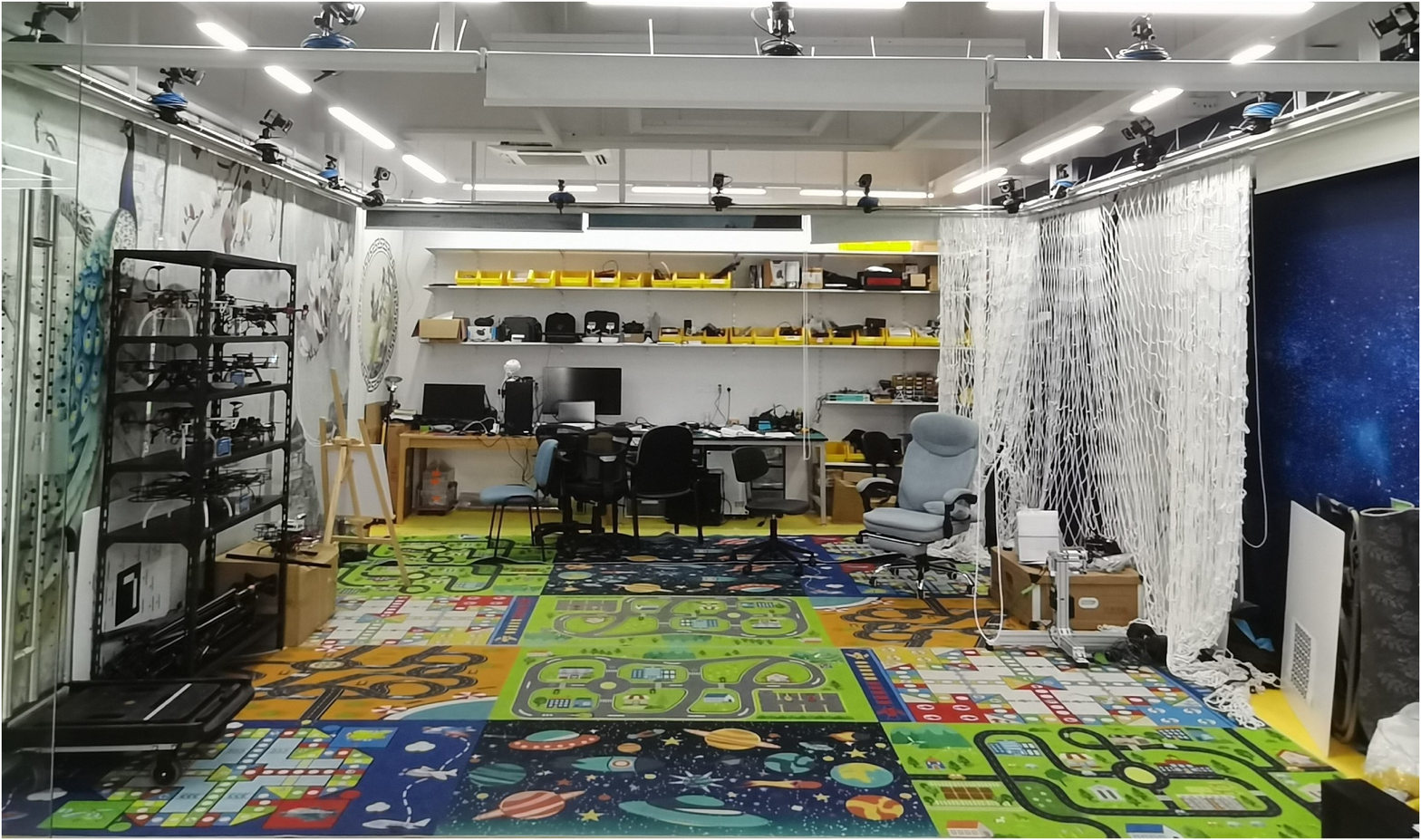} &
				\includegraphics[scale=0.07128]{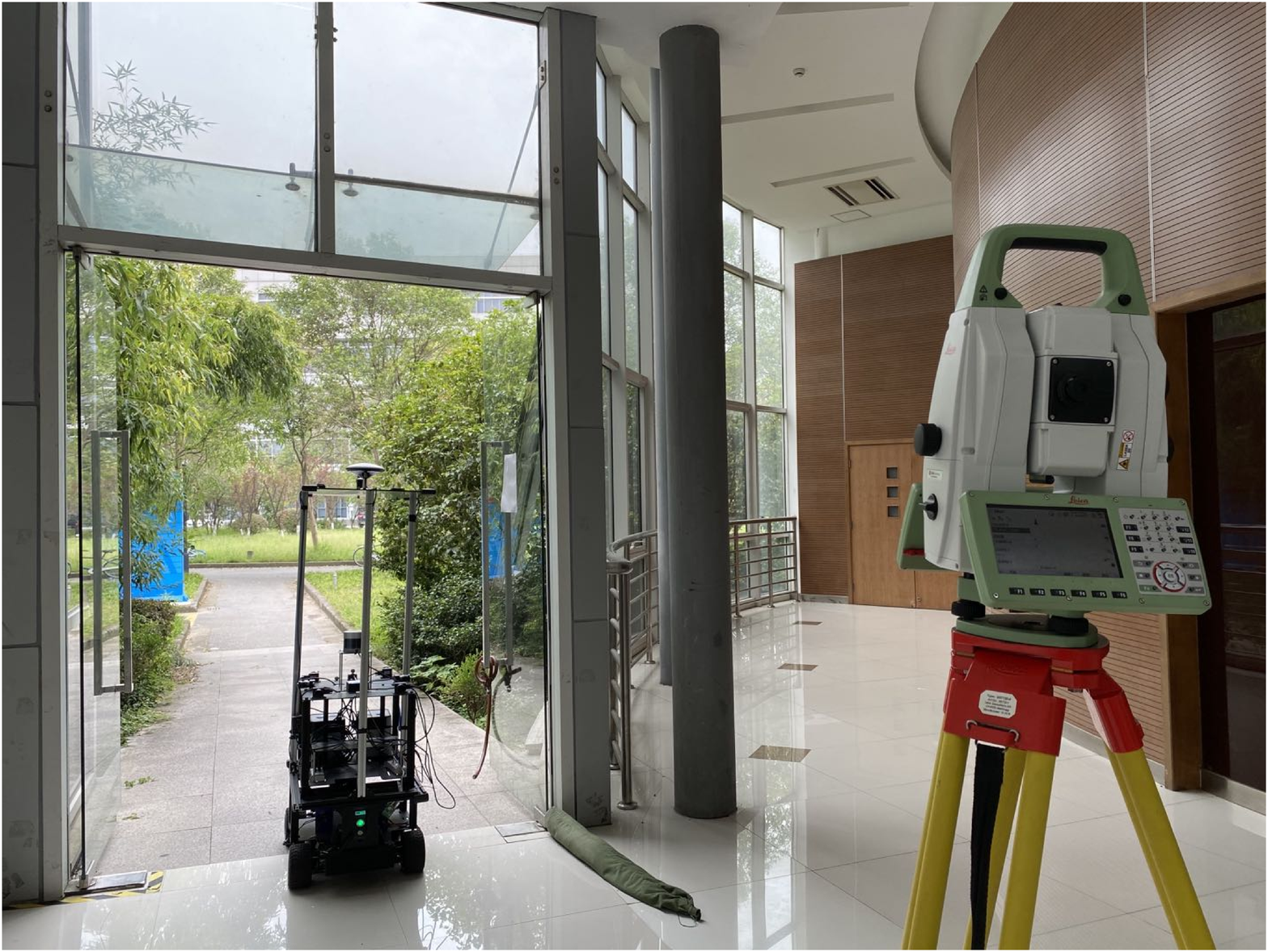}&
				\includegraphics[scale=0.07128]{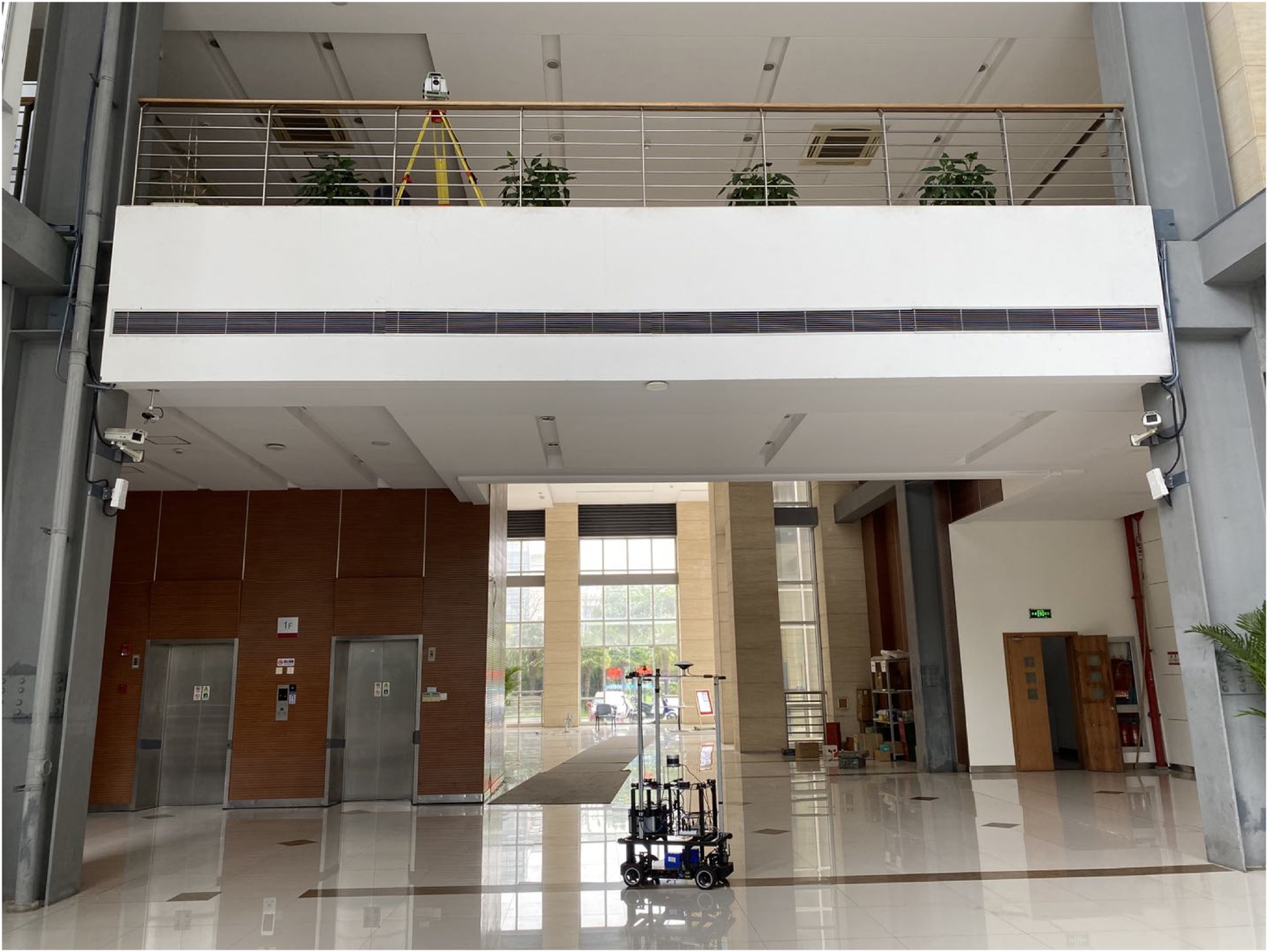}
				
				\\
				(a) & (b) & (c) & (d)\\
				
			\end{tabular}
		\end{center}
		\caption{(a) We visualize the trajectories of outdoor sequences in the map with different colors. (b) The ground truth of indoor sequences are acquired from a motion-capture system with twelve cameras. (c) A 3D laser tracker is used to track the robot through a door from indoors to outdoors for door sequences. (d) In lift sequences, the robot moved in a hall on the first floor and then went to the second floor by lift. A 3D laser tracker is used to track its ground-truth trajectory outside the lift. }
		\label{scenarios}
	\end{figure*}
	\paragraph{Datasets with event Camera}
	E. Mueggler et al. \cite{mueggler2017event} used the event camera to collect events of dynamic and static scenes inside and outside the room. The recording time of this dataset is too short to evaluate the performance of SLAM algorithms reliably. Interiornet dataset \cite{li2018interiornet} simulates sensors including RGBD, IMU, stereo cameras, and event cameras, but there is a gap between the simulated and the real-world scenes.
	\begin{table*}[h]
		\centering
		\caption{Specifications of sensors and tracking devices}
		\label{sensor}
		\begin{tabular}{cccc}
			\hline
			Device & Type  & Spec. & Freq.(Hz) \\
			\hline
			LiDAR & Velodyne VLP-32C & 32 beam,360 H-FOV,40V-FOV & 10 \\
			IMU   & Handsfree A9 & 9-axis & 150 \\
			GNSS Receiver & Ublox M8T & GPS/BeiDou & 1 \\
			RGB Camera & FLIR Pointgrey CM3-U3-13Y3C & 1280*1024, 190 H-FOV, 190 V-FOV & 15 \\
			Infrared Camera & Gaode PLUG 617 & 640*512, 90.2 H-FOV,70.6 V-FOV & 25 \\
			Event Camera & Inivation DVXplorer & 640*480, 65.2 H-FOV,51.3 V-FOV & 15 \\
			\multirow{2}[0]{*}{VI-sensor} & \multirow{2}[0]{*}{Realsense d435i} & RGB: 640*480, 69 H-FOV, 42.5V-FOV & 15 \\
			&       & IMU: 6-axis & 200 \\
			\hline
			Mocap System & Vicon Vero 2.2 & localization accuracy 1mm & 50 \\
			Laser Tracker & Leica Nova MS60 & localization accuracy 1mm + 1.5ppm & 10 \\
			\multirow{2}[0]{*}{RTK/INS} & \multirow{2}[0]{*}{Xsens Mti 680G} & RTK: localization accuracy 2cm & 100 \\
			&       & INS: 9-axis & 100 \\
			\hline
		\end{tabular}%
		\label{tab:addlabel}%
	\end{table*}%
	
	\paragraph{SLAM datasets with GNSS}
	Currently, there are very few SLAM datasets containing raw GNSS information. UrbanLoco dataset \cite{wen2020urbanloco} collected the raw GNSS data with Ublox M8T and used a fish-eye camera to capture the sky, but it did not publish the data of the camera. As far as we know, there is no public SLAM dataset containing both GNSS raw measurements and images of a sky-pointing camera, while such kind of data could be valuable to some research. 
	
	In summary, we review the current best-known SLAM datasets and analyze their limitations. Table \ref{dataset comparison} summarizes the defects and limitations of some mentioned datasets. From the table, we conclude that currently a dataset with rich scenarios and complete sensor information is urgently needed for furthering the study of SLAM algorithms for ground robots. To fill this gap,  we present a new dataset M2DGR.
	Compared with previous SLAM datasets, our dataset contains long-term trajectories in diverse real-world scenarios with a rich pool of sensory information, which facilitates tests and comparisons of various algorithm designs.

	\section{THE M2DGR DATASET}
	\subsection{Acquisition platform}	
	We construct a ground robot for data collection as shown in Figure \ref{car}. This robot has three layers. The bottom layer contains the power supply, the computer, and the display. The middle layer and the top layer include different sensors. The dimension figures of our robot are shown in Figure \ref{car} (a) $\sim$ (c).
	To ensure the high-speed transmission of data, we connect the LiDAR to the Ethernet port of the host and other sensory devices to the USB3.0 port of the host. We record the data on a high-end laptop with a high-speed NMVe SSD.

	\subsection{Sensor setup}
	The location of all the devices mounted is shown in Figure \ref{car}. Six fish-eye cameras were used to capture the images of the surroundings in $360^\circ$ field of view, including forward-looking stereo cameras, rear-looking stereo cameras, and two side-looking cameras. Additionally, a $32$-beam Velodyne LiDAR was used to scan the surrounding environment and obtain the 3D point cloud. 
	We also used an infrared camera to capture infrared thermal images, a VI-sensor to obtain forward-looking color images as well as inertial data, and an event camera to capture dynamic information. We also mounted a consumer-level IMU, a GNSS receiver to collect GNSS raw signals, and a sky-pointing fish-eye camera to monitor the sky. For ground truth of trajectories, we applied Xsens Mti 680G, a GNSS-IMU integrated navigation system, to track the robots in outdoor sequences, while used a motion-capture system and a laser scanner to track the robot in indoor sequences. All the sensors and tracking devices as well as their key parameters are listed in Table \ref{sensor}.

	\begin{table*}[h]
		\caption{An overview of scenarios in our dataset}
		\label{overview}
		\centering
		\begin{tabular}{p{1.5cm}p{1.25cm}p{1.25cm}p{1.25cm}p{1.25cm}p{0.95cm}p{0.95cm}p{0.95cm}p{0.9cm}p{1.4cm}p{1.75cm}p{1.3cm}}
			\hline
			\makecell[c]{Scenario} & \makecell[c]{Street} & \makecell[c]{Circle}  & \makecell[c]{Gate}  & \makecell[c]{Walk} & \makecell[c]{Hall} & \makecell[c]{Door}  & \makecell[c]{Lift} & \makecell[c]{Room} & \makecell[c]{Roomdark}& \makecell[c]{TOTAL}\\
			\hline
			\makecell[c]{Number} &\makecell[c]{10} & \makecell[c]{2}  & \makecell[c]{3}  & \makecell[c]{1}  & \makecell[c]{5} & \makecell[c]{2} & \makecell[c]{4} & \makecell[c]{3} & \makecell[c]{6} & \makecell[c]{36}\\
			
			\makecell[c]{Size/GB} & \makecell[c]{590.7}  & \makecell[c]{50.6}  & \makecell[c]{65.9}  & \makecell[c]{21.5} & \makecell[c]{117.4} & \makecell[c]{46.0} & \makecell[c]{112.1} & \makecell[c]{45.3} & \makecell[c]{171.1}& \makecell[c]{1220.6}\\
			
			\makecell[c]{Duration/s} & \makecell[c]{7958}  & \makecell[c]{478}  & \makecell[c]{782}  & \makecell[c]{291} & \makecell[c]{1226} & \makecell[c]{588} & \makecell[c]{1224} & \makecell[c]{275} & \makecell[c]{866}& \makecell[c]{13688}\\

			\makecell[c]{Dist/m} & \makecell[c]{7727.72}  & \makecell[c]{618.03}  & \makecell[c]{248.40}  & \makecell[c]{263.17} & \makecell[c]{845.15} & \makecell[c]{200.14} & \makecell[c]{266.27} & \makecell[c]{144.13} & \makecell[c]{395.66}& \makecell[c]{10708.67}\\

			\makecell[c]{GT} & \makecell[c]{RTk/INS}  & \makecell[c]{RTk/INS}  & \makecell[c]{RTk/INS}  & \makecell[c]{RTk/INS} & \makecell[c]{Leica} & \makecell[c]{Leica} & \makecell[c]{Leica} & \makecell[c]{Mocap} & \makecell[c]{Mocap}& \makecell[c]{---}\\

			\hline     
			
		\end{tabular}
	\end{table*}

	\begin{table*}[h]
		\caption{Sample sequences for evaluation}
		\label{sequence feature}
		\centering
		\begin{tabular}{p{1.7cm}p{1.5cm}p{1.7cm}p{1.7cm}p{2.4cm}p{1.7cm}p{1.6cm}p{2.6cm}p{2.8cm}}
			\hline
			\makecell[c]{Sequence} & \makecell[c]{Street02} & \makecell[c]{Street06}  & \makecell[c]{Street07}  & \makecell[c]{Roomdark06} & \makecell[c]{Hall05} & \makecell[c]{Door01}  & \makecell[c]{Lift04} \\
			\hline
			\makecell[c]{Duration/s} & \makecell[c]{1227}  & \makecell[c]{494}  & \makecell[c]{929}  & \makecell[c]{172} & \makecell[c]{402} & \makecell[c]{461} & \makecell[c]{299} \\

			\makecell[c]{Distance/m} & \makecell[c]{1484.62} & \makecell[c]{479.63} & \makecell[c]{1104.07}& \makecell[c]{72.53}  & \makecell[c]{79.28}  & \makecell[c]{285.51}  & \makecell[c]{142.78}  \\

			\makecell[c]{Speed/(m/s)} & \makecell[c]{1.21}  & \makecell[c]{0.97}  & \makecell[c]{1.19}  & \makecell[c]{0.42} & \makecell[c]{0.71} & \makecell[c]{0.31} & \makecell[c]{0.27} \\
			\hline
			
			\makecell[c]{Description} & \makecell[c]{day,}  & \makecell[c]{night,}  & \makecell[c]{night,}  & \makecell[c]{room,} & \makecell[c]{long-term} & \makecell[c]{outdoors to} & \makecell[c]{first floor to second} \\
			
			\makecell[c]{of features} & \makecell[c]{long-term,}  & \makecell[c]{straight line}  & \makecell[c]{zigzag route}  & \makecell[c]{complete darkness} & \makecell[c]{large overlap} & \makecell[c]{outdoors} & \makecell[c]{floor by lift} \\
			
			\hline     
			
		\end{tabular}
	\end{table*}

		


		


	\begin{table*}[h]
		\caption{ATE(m) of SLAM systems on sample sequences}
		\label{ate rmse tab}
		\centering
		\begin{tabular}{p{2.8cm}p{1.8cm}p{1.3cm}p{1.3cm}p{1.8cm}p{1.3cm}p{1.3cm}p{1.3cm}p{1.3cm}}
			\hline
			\makecell[c]{Method / Sequence} & \makecell[c]{Street02} & \makecell[c]{Street06}  & \makecell[c]{Street07}  & \makecell[c]{Roomdark06} & \makecell[c]{Hall05} & \makecell[c]{Door01}  & \makecell[c]{Lift04} \\
			\hline

			\makecell[c]{A-LOAM \cite{zhang2014loam}} & \makecell[c]{5.299}  & \makecell[c]{0.628}  & \makecell[c]{28.940}  & \makecell[c]{0.314} & \makecell[c]{1.065} & \makecell[c]{0.274} & \makecell[c]{1.323} \\
			
			\makecell[c]{LeGO-LOAM \cite{shan2018lego}} & \makecell[c]{20.021}  & \makecell[c]{1.246}  & \makecell[c]{35.437}  & \makecell[c]{0.373} & \makecell[c]{1.030} & \makecell[c]{0.253} & \makecell[c]{1.370} \\
			
			\makecell[c]{LINS \cite{qin2020lins}} & \makecell[c]{5.636}  & \makecell[c]{1.742}  & \makecell[c]{12.009}  & \makecell[c]{2.205} & \makecell[c]{1.010} & \makecell[c]{0.258} & \makecell[c]{1.318} \\

			\makecell[c]{LIO-SAM \cite{shan2020lio}} & \makecell[c]{4.063}  & \makecell[c]{0.417}  & \makecell[c]{28.642}  & \makecell[c]{0.324} & \makecell[c]{1.047} & \makecell[c]{0.268} & \makecell[c]{1247.153} \\
			\hline 
			\makecell[c]{ORB3-Pinhole \cite{campos2021orb}} & \makecell[c]{152.462}  & \makecell[c]{5.845}  & \makecell[c]{X$^{\mathrm{a}}$}  & \makecell[c]{X} & \makecell[c]{3.291} & \makecell[c]{7.662} & \makecell[c]{X} \\

			\makecell[c]{ORB3-Fisheye \cite{campos2021orb}} & \makecell[c]{X}  & \makecell[c]{95.056}  & \makecell[c]{X}  & \makecell[c]{X} & \makecell[c]{X} & \makecell[c]{2.295} & \makecell[c]{8.131} \\

			\makecell[c]{ORB3-Thermal \cite{campos2021orb}} & \makecell[c]{154.778}  & \makecell[c]{30.450}  & \makecell[c]{8.863}  & \makecell[c]{0.404} & \makecell[c]{5.927} & \makecell[c]{1.241} & \makecell[c]{2.873} \\
			
			\makecell[c]{CubemapSLAM \cite{wang2018cubemapslam}} & \makecell[c]{X}  & \makecell[c]{98.391}  & \makecell[c]{X}  & \makecell[c]{X} & \makecell[c]{5.171} & \makecell[c]{9.328} & \makecell[c]{X} \\

			\makecell[c]{VINS-Mono \cite{qin2018vins}} & \makecell[c]{24.157}  & \makecell[c]{124.357}  & \makecell[c]{143.725}  & \makecell[c]{1.001} & \makecell[c]{0.646} & \makecell[c]{0.694} & \makecell[c]{5.582} \\
			\hline 
			\makecell[c]{RTKLIB \cite{takasu2009development}} & \makecell[c]{7.072}  & \makecell[c]{6.749}  & \makecell[c]{13.096}  & \makecell[c]{X} & \makecell[c]{X} & \makecell[c]{5.344} & \makecell[c]{X} \\
			\hline
			\multicolumn{8}{l}{$^{\mathrm{a}}$If a visual SLAM fails to initialize or track frames less than a half of total frames or a GNSS-based method fails to initialize, we mark it X}
			
		\end{tabular}
	\end{table*}
	
	\subsection{Calibration and synchronization}
	We use the MATLAB camera calibration toolbox to obtain the camera intrinsics of pinhole cameras. For fish-eye cameras, we use Kannala Brandt model \cite{kannala2006generic}, Omnidirectional model \cite{scaramuzza2006toolbox},  and MEI model \cite{mei2007single}  for calibration. To calibrate thermal-infrared cameras, we heat a checkerboard where black blocks and white blocks were made of materials with different heat capacities.  Thus, those blocks can be identified by the infrared camera easily. We use toolbox \cite{UCAM-CL-TR-696} to calibrate internal parameters of IMU, including the white noise and random walk of both the gyroscopic and the accelerometer measurements. 
	
	We choose the LiDAR frame as the reference to calibrate the extrinsic parameters (relative poses) between sensors. We use the toolbox \cite{lv2020targetless} to calibrate the extrinsic parameters between IMU and LiDAR,  and Kalibr toolbox \cite{furgale2013unified} to calibrate the extrinsic parameters between cameras and IMU, as well as Autoware software \cite{kato2015open} to calibrate extrinsic parameters between LiDAR and cameras. 
	
	We do not use hardware signals to trigger all the sensors to capture data at the same time but record the data from different sensors using the same system time stamps. The cameras, including the six fisheye cameras and the sky-pointing camera, are synchronized by software  - triggering data capturing at the same instance by calling the API. Our tests show that such a soft synchronization approach achieves accurate time synchronization of less than $10$ ms between different cameras.
	
	\subsection{Data collection}	
	We operated our robot traveling around various scenes. An overview of our dataset is given in Table \ref{overview}. The recording processes are described as follows:

	For outdoor environments, we collected sequences on the campus of Shanghai Jiao Tong University. The satellite visibility was good so that the GNSS-RTK suite outputs high-accuracy ground truth of trajectories. To better test loop closing of visual SLAM, we particularly recorded sequences \emph{Circle01} and \emph{Circle02} while the robot was circling in repeated routes. All the outdoor trajectories are visualized in Figure \ref{scenarios} (a).
	
\begin{figure}[h]
		\centering
		\includegraphics[width=0.44\linewidth]{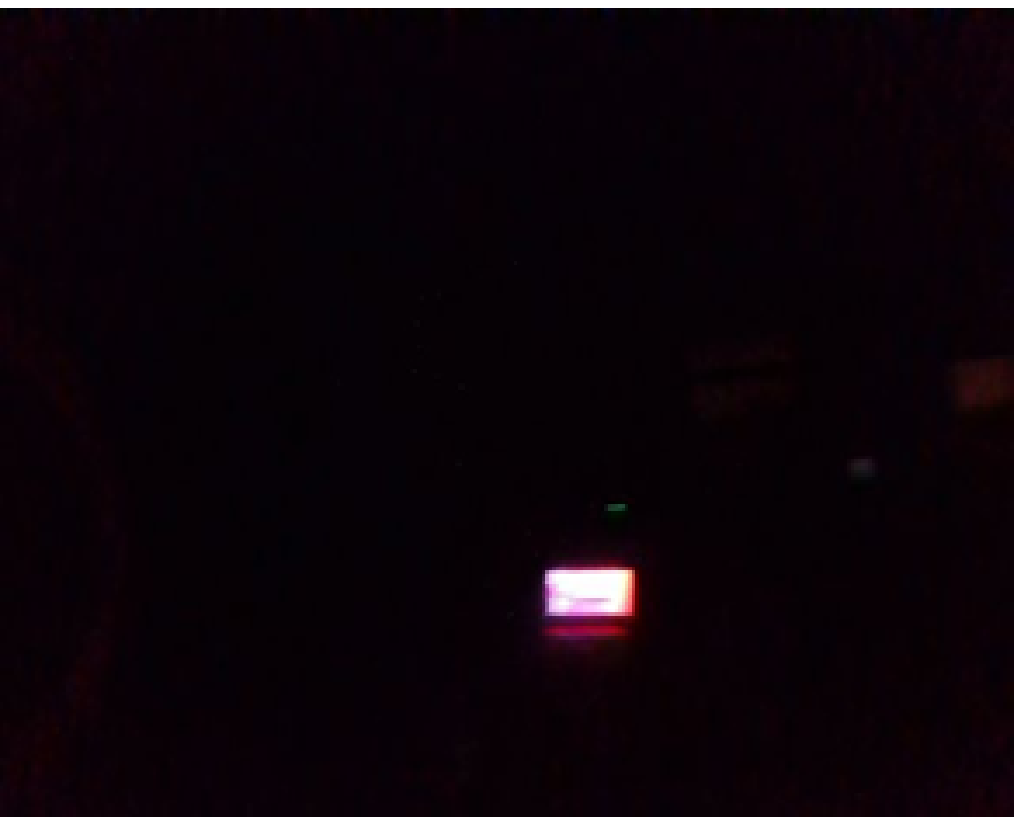}
		\includegraphics[width=0.44\linewidth]{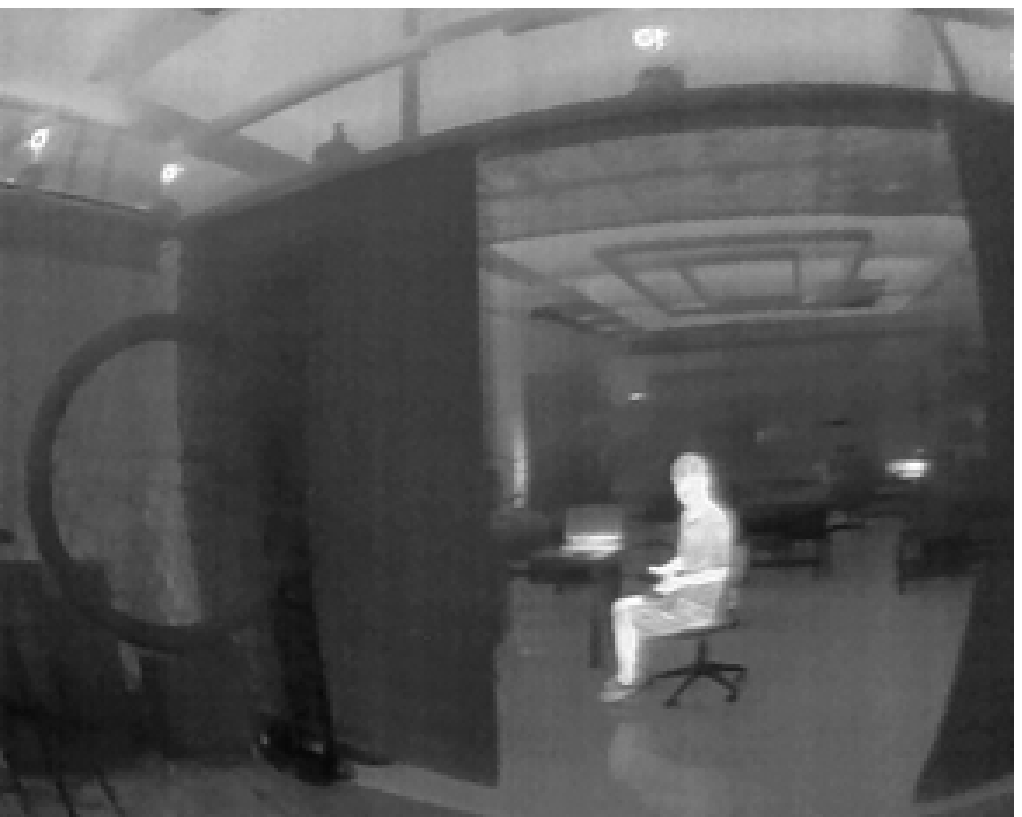}
		
		\caption{Color image (\textbf{Left}) and Infra-red image (\textbf{Right}) in a complete dark scene.}
		\label{darkandlift}
	\end{figure}

	For indoor environments, we obtained the ground truth trajectories with a motion-capture system  with twelve high-speed
	tracking cameras (50 Hz) in a room, as illustrated in Figure \ref{scenarios} (b). Particularly, we recorded a few sequences in complete darkness to test SLAM systems' robustness, as shown in Figure \ref{darkandlift}.
	Outside the room with the motion-capture system, we used a laser tracker to generate the ground truth of trajectories indoors. A prism reflector was mounted on our robot so that the laser tracker could well track it. Also, to better evaluate loop closing of visual SLAM, we recorded the sequence \emph{Hall05} where the robot circled around a hall in repeated routes. 
	
	We also recorded sequences in a mixture of indoor and outdoor environments. We operated the robot traveling outdoors for some time until GNSS signals were well received. Then we got the robot entering a hall through a door. After the robot moved around the hall for a period, we got it going outdoors again. Those sequences are used to evaluate the performance of SLAM and GNSS positioning methods on the critical point between indoors and outdoors.
	
	Lastly, we collected sequences to test the capability of entering and leaving the lift. More specifically, we manipulated the robot to travel around within a hall on the first floor and then enter a lift that carried the robot to the second floor, as Fig \ref{scenarios} (d) shows.

	\subsection{Data usage and tools}

	All the data were captured by rosbag in Robot Operation System (ROS), and the recorded topics are listed as follows.

	\begin{itemize}
		\item RGB camera: \\
		/camera/left/image$\_$raw/compressed\\
		/camera/right/image$\_$raw/compressed\\ 
		/camera/third/image$\_$raw/compressed\\
		/camera/fourth/image$\_$raw/compressed\\
		/camera/fifth/image$\_$raw/compressed\\
		/camera/sixth/image$\_$raw/compressed\\
		/camera/head/image$\_$raw/compressed
		\item VI-sensor:\\
		/camera/color/image$\_$raw/compressed \\/camera/imu
		\item Raw GNSS:\\ 
		/ublox/aidalm\\
		/ublox/rxmraw\\
		/ublox/fix\\
		/ublox/navstatus
		\item Event camera: \\
		/dvs$\_$rendering/compressed\\/dvs/events0
		\item LiDAR: \\
		/velodyne$\_$points
		\item Infrared Camera:\\
		/thermal$\_$image$\_$raw
		
		\item IMU:\\
		/handsfree/imu
		
	\end{itemize}
	
	For convenience, we provide scripts to export the data to other formats such as \cite{burri2016euroc}. Ground-truth trajectories and calibration results are provided for each sequence. Furthermore, we give detailed instructions to evaluate the performance of different SLAM algorithms on our project page.

	\section{Evaluation}
	\begin{figure*}[h]
		\begin{center}
			\footnotesize
			\begin{tabular}{cccc}
				\includegraphics[scale=0.29]{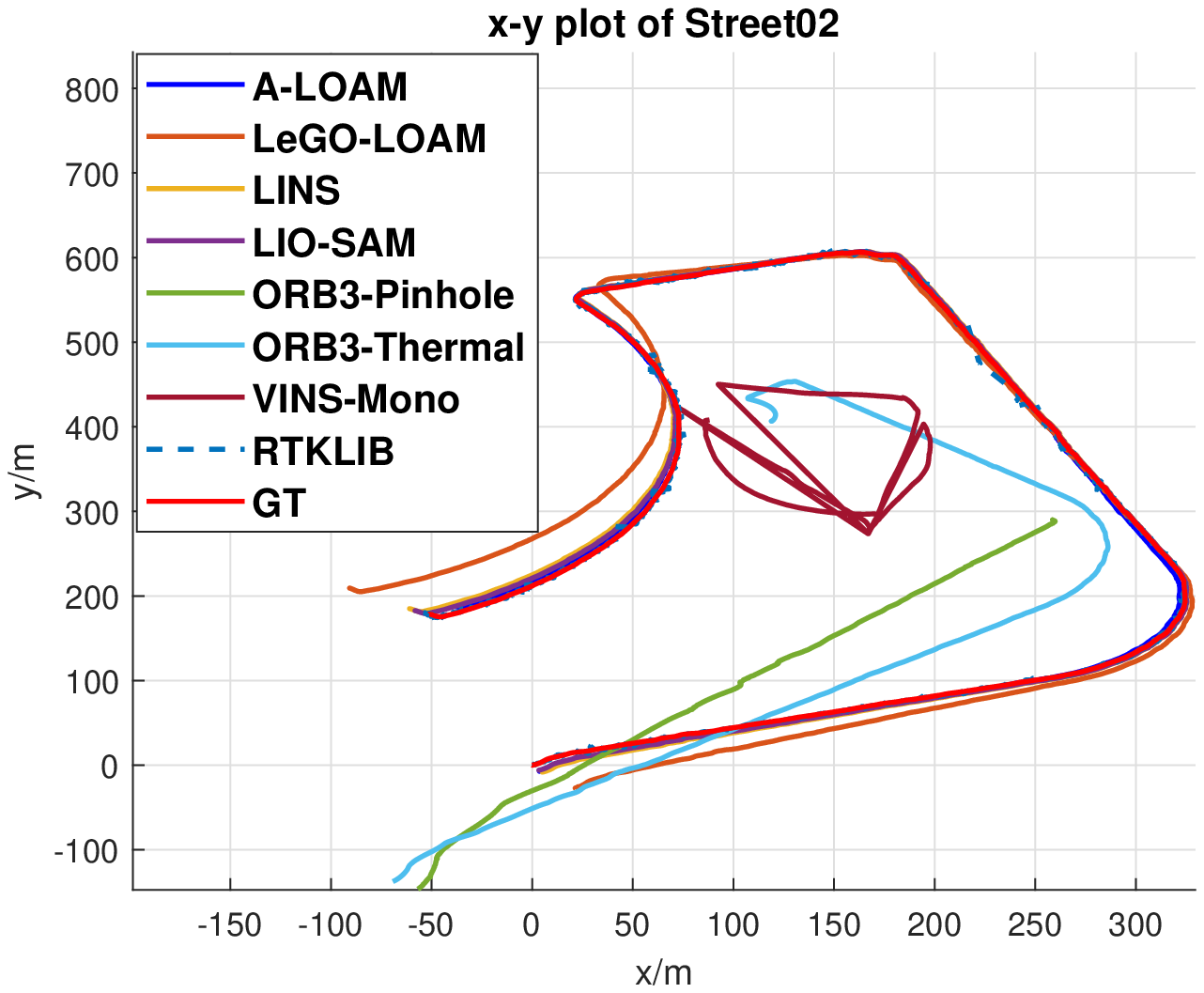} &
				\includegraphics[scale=0.29]{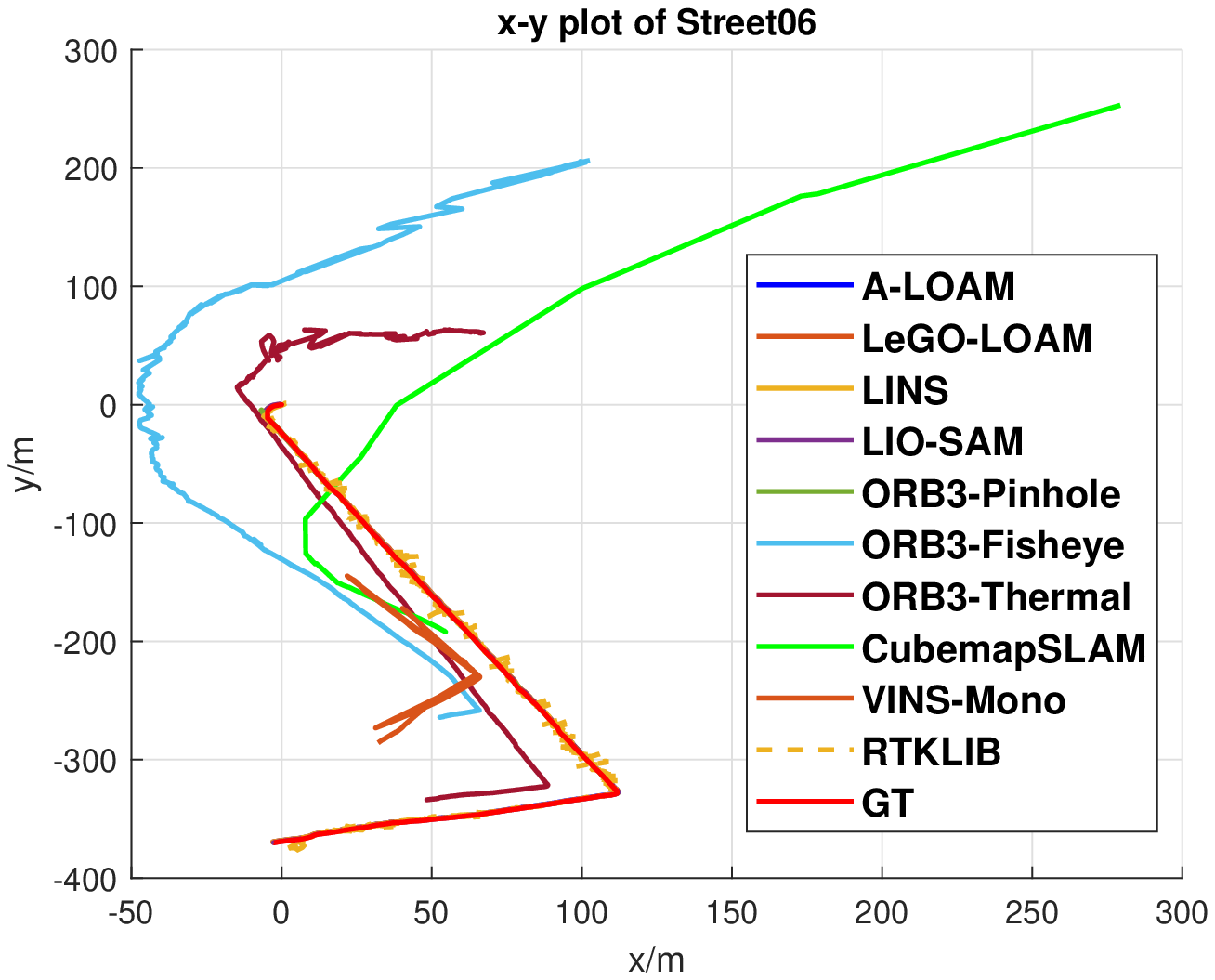} &
				\includegraphics[scale=0.29]{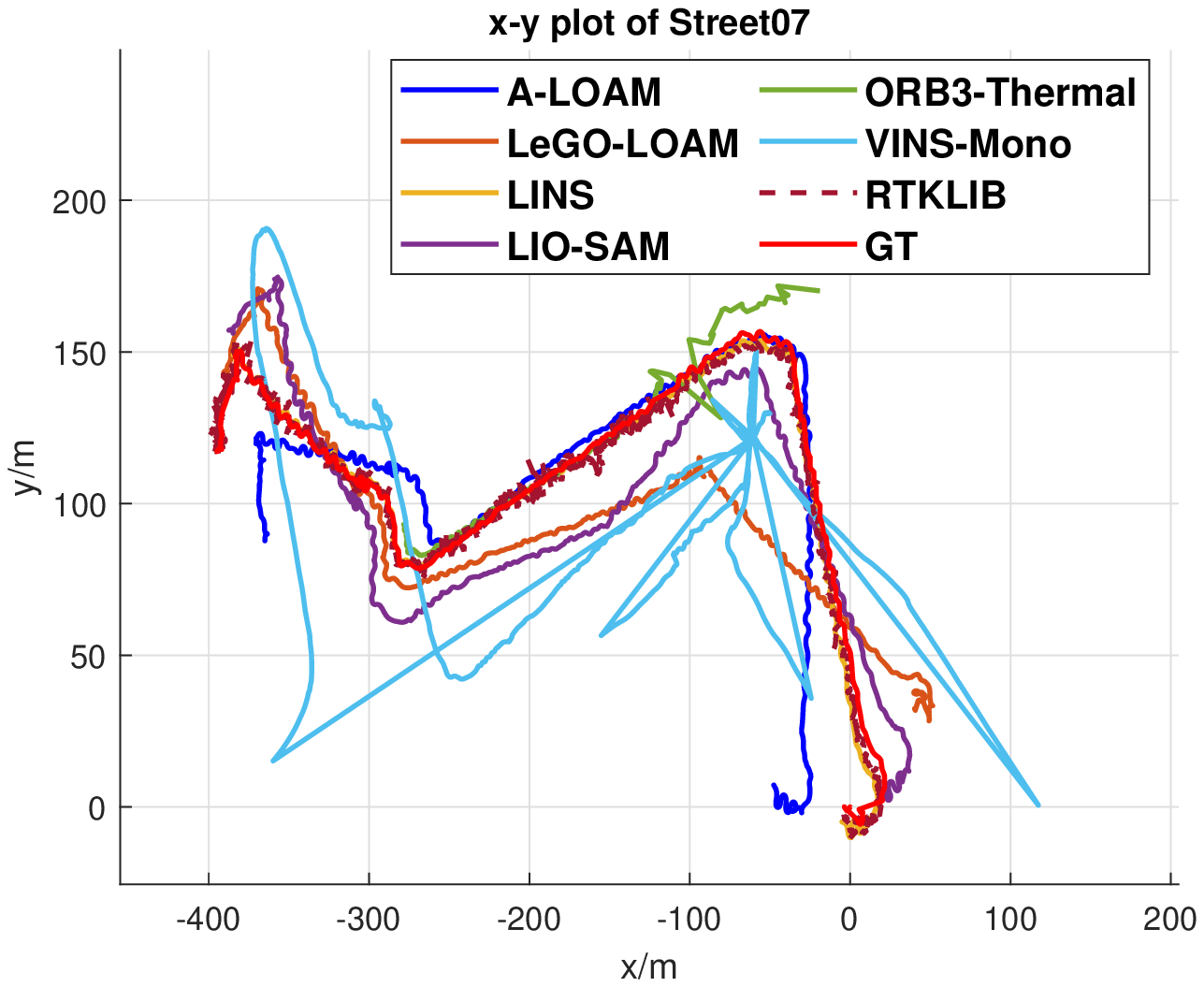}&
				\includegraphics[scale=0.29]{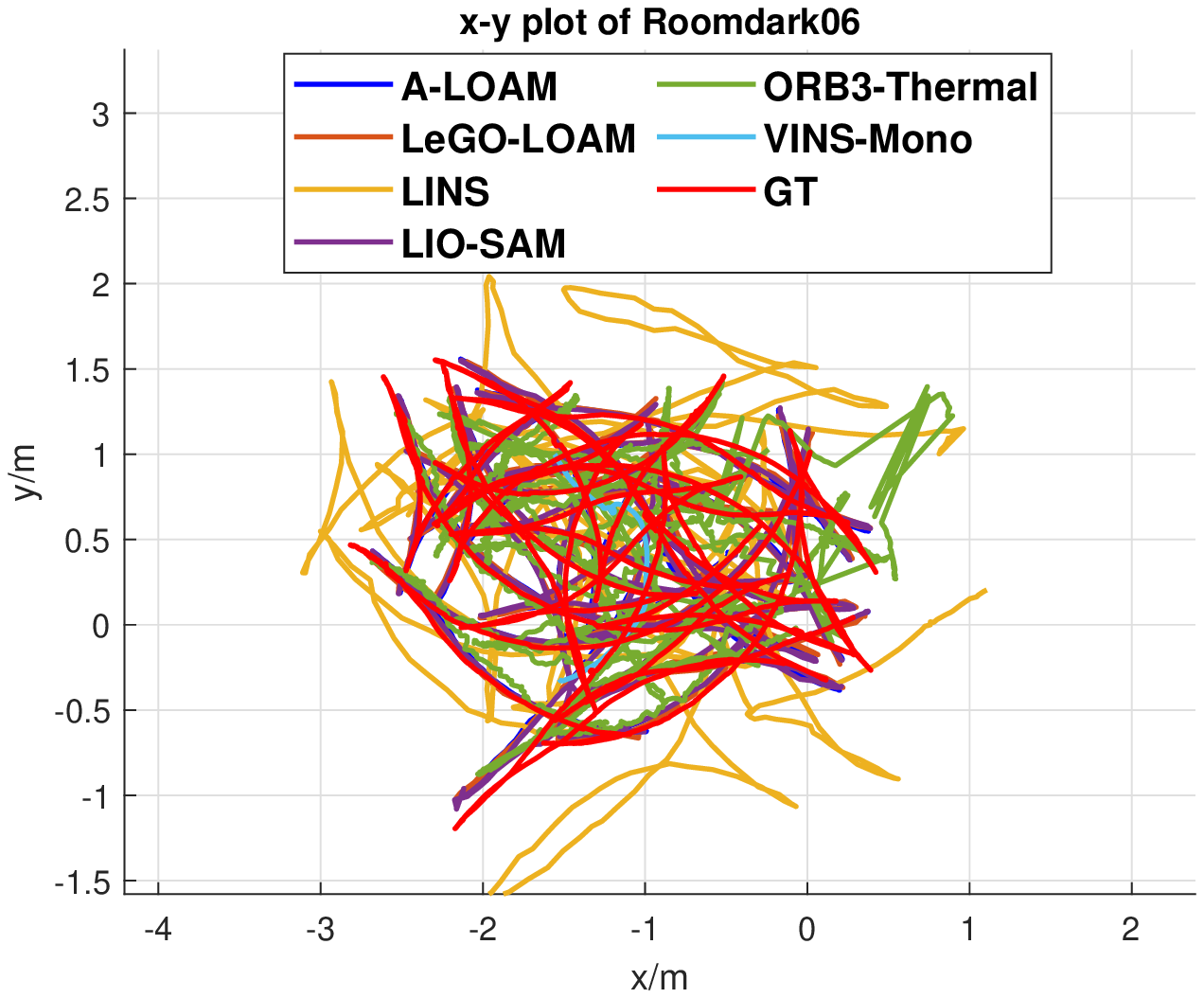}\\
				(a) x-y graph of Street02 & (b) x-y graph of Street06& (c) x-y graph of Street07& (d) x-y graph of Roomdark06\\
				
				\includegraphics[scale=0.29]{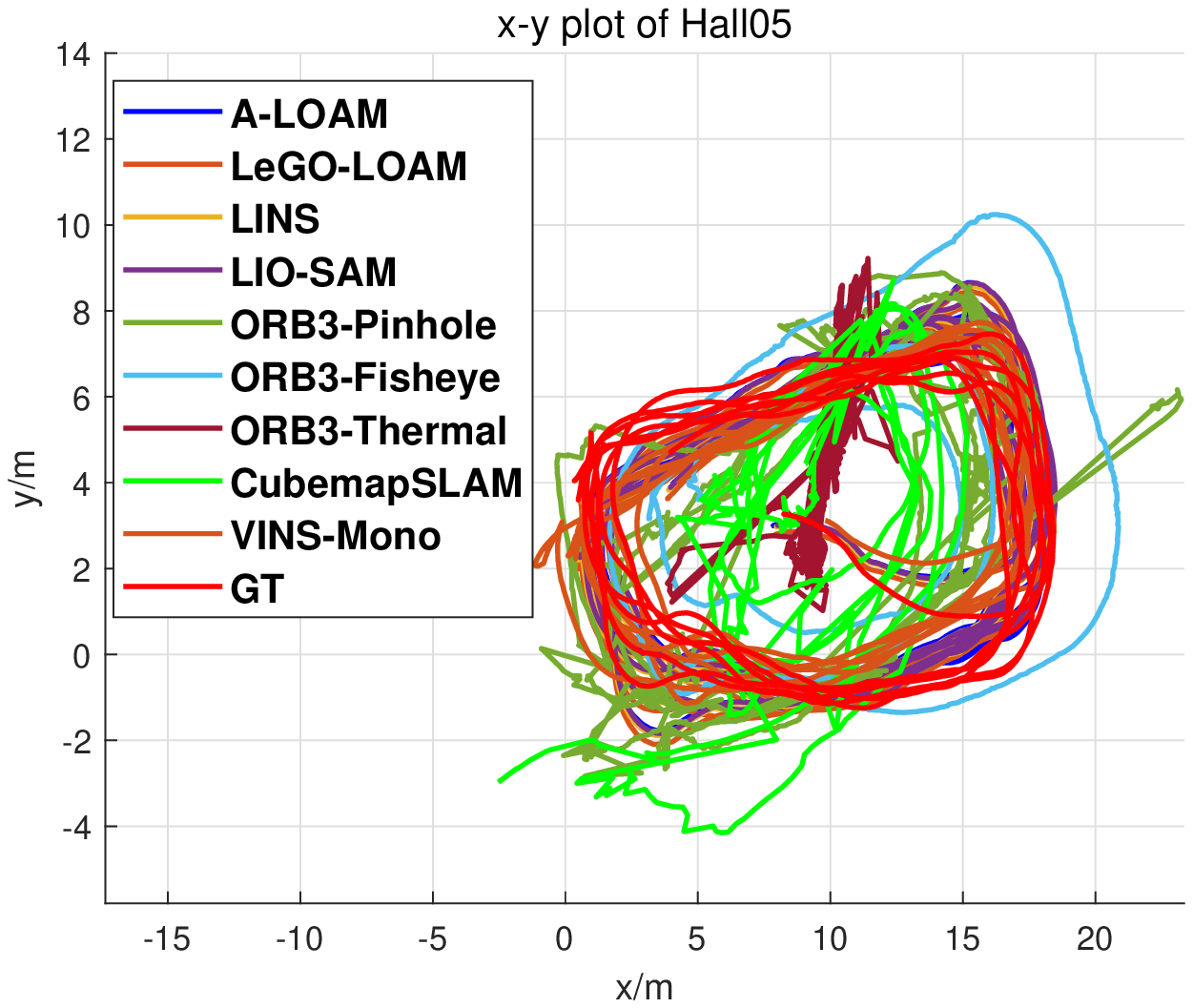}&
				\includegraphics[scale=0.29]{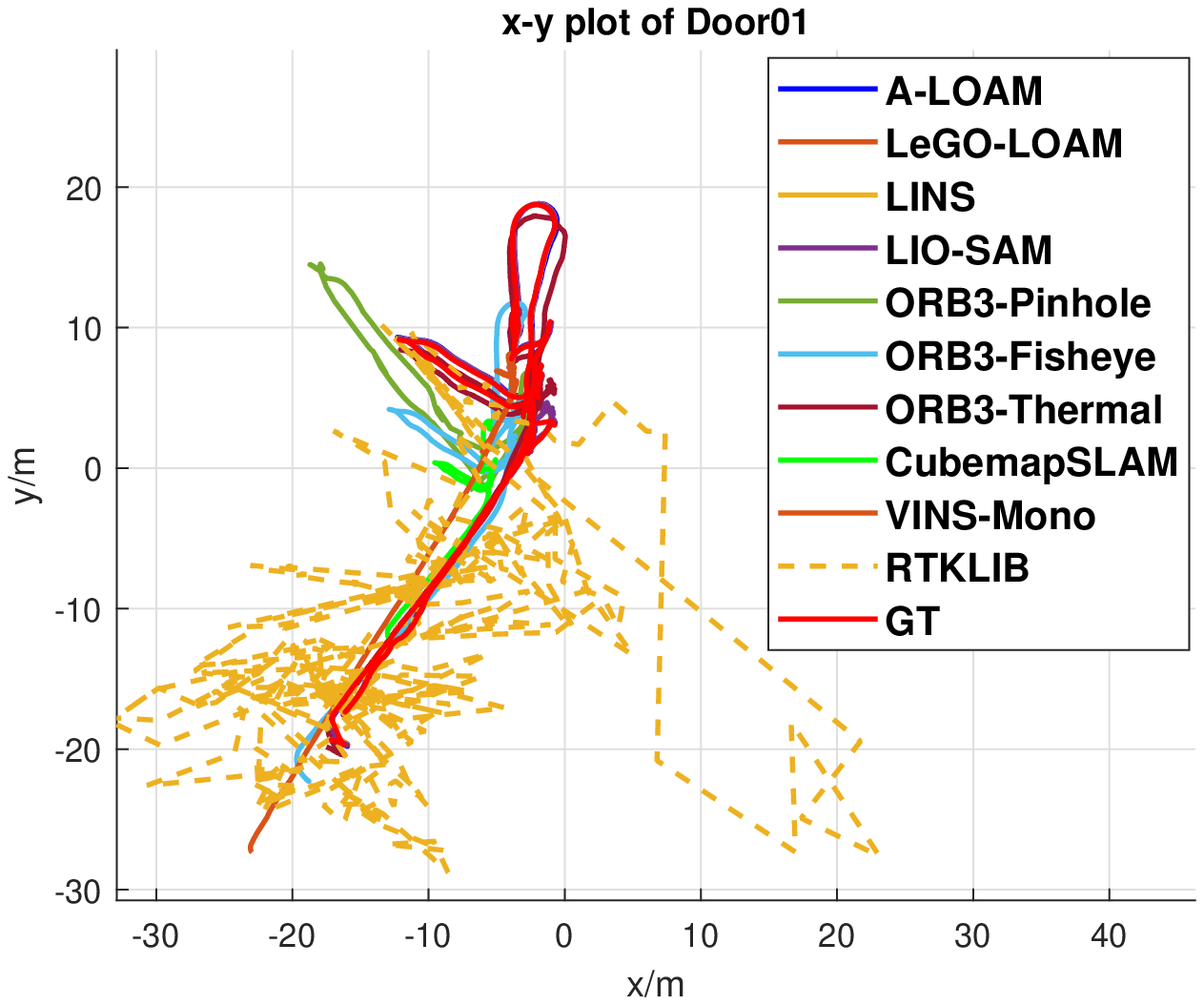}&
				\includegraphics[scale=0.29]{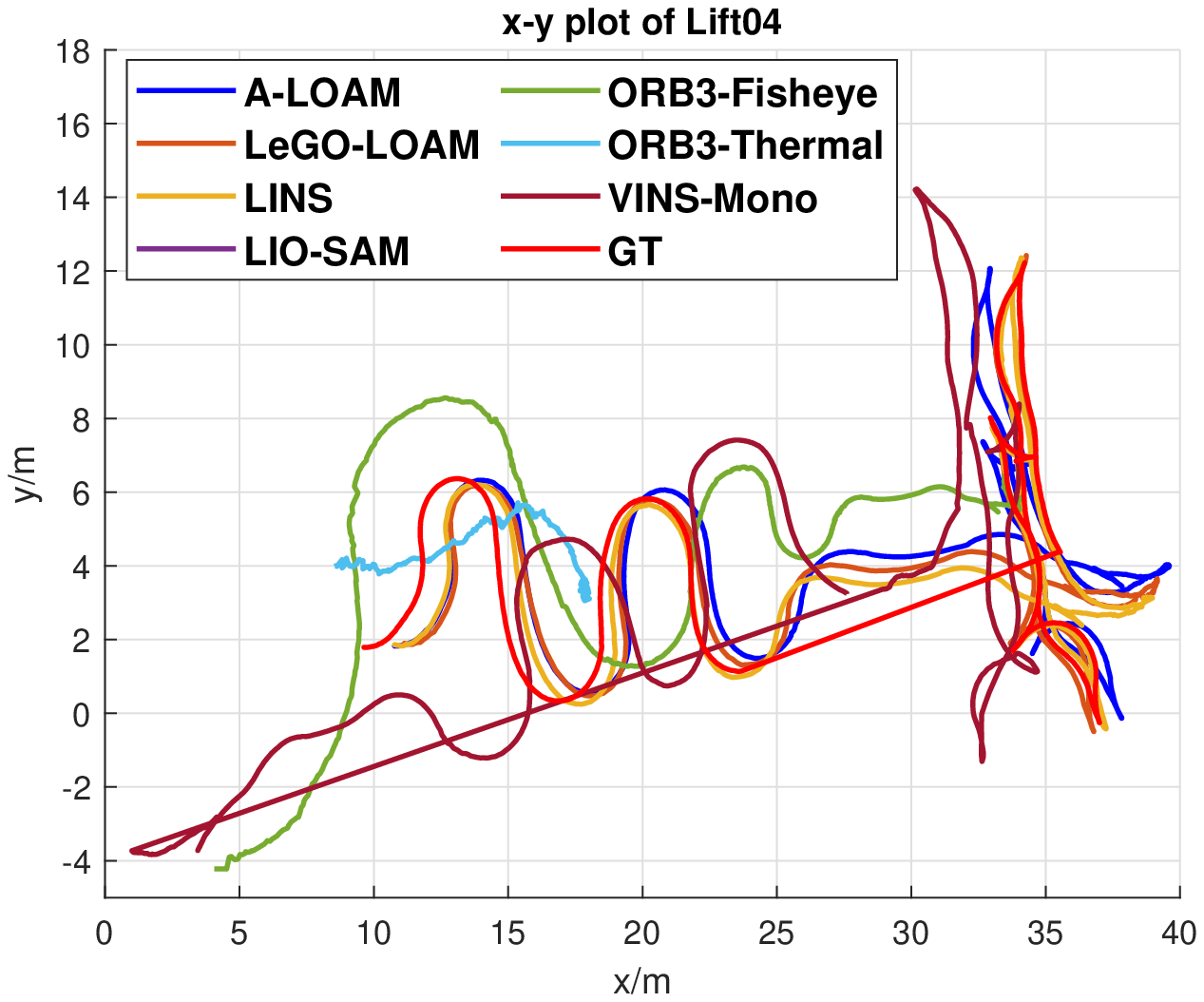}&
				\includegraphics[scale=0.29]{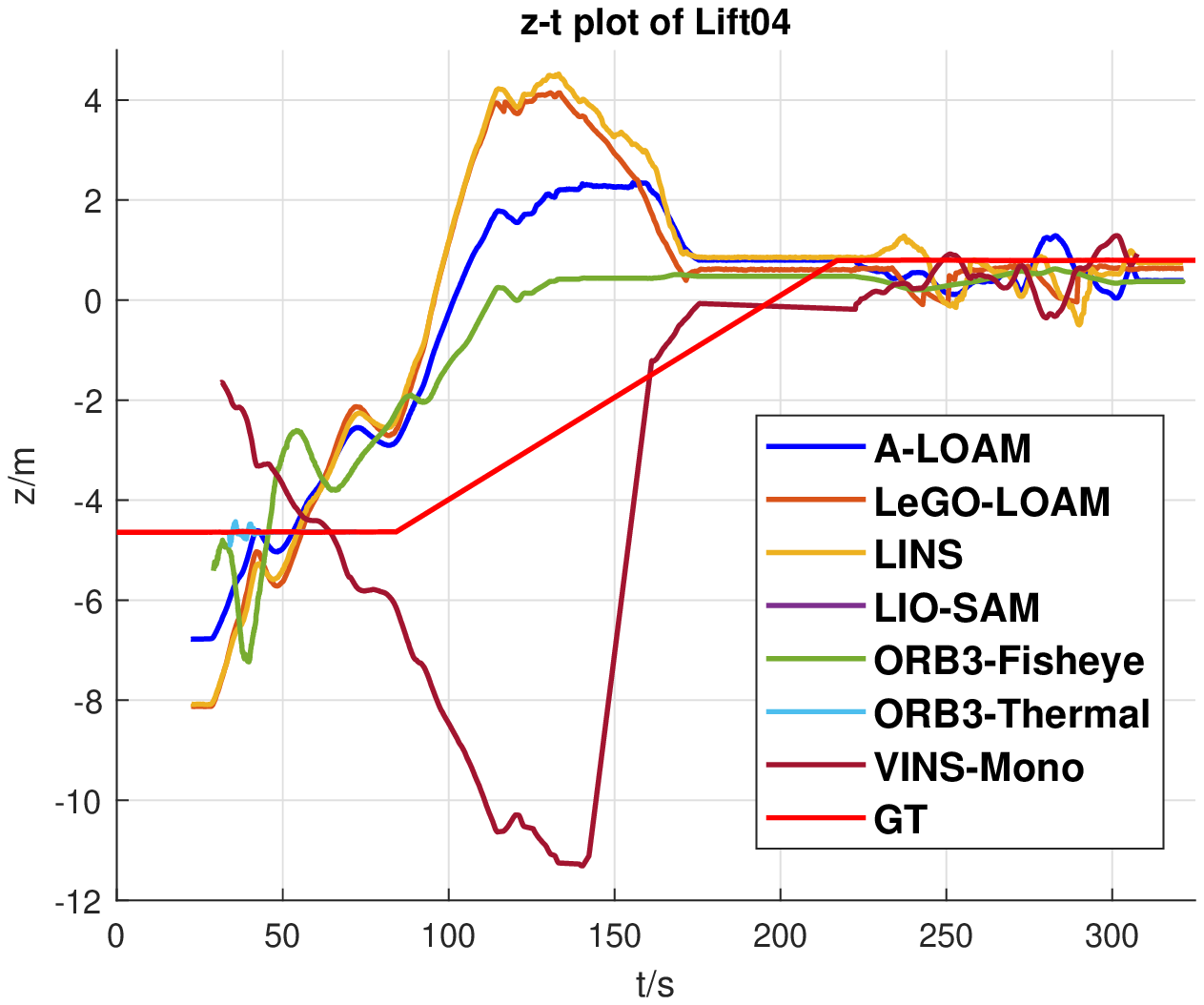}\\
				(e) x-y graph of Hall05& (f) x-y graph of Door01& (g) x-y graph of Lift04 & (h) time-height graph of Lift04\\
			\end{tabular}
		\end{center}
		\caption{Estimated and ground-truth (GT) trajectories of $7$ sample sequences are visualized in ENU (East-North-Up) coordinate system. }
		\label{ate rmse fig}
	\end{figure*}

				
				
	
	We evaluate the state-of-the-art SLAM systems on seven representative sequences from our dataset. Those sequences are described in Table \ref{sequence feature}. 
	The Absolute Trajectory Error (ATE) \cite{sturm2012benchmark} is used for the evaluation metric. All the estimated trajectories are aligned with the ground truth by the EVO tool \cite{MichaelGrupp2018} to obtain the ATE errors. 
	
	For visual SLAM, we test ORB-SLAM3 \cite{campos2021orb}, CubemapSLAM \cite{wang2018cubemapslam}, Multicol-SLAM \cite{urban2016multicol}, and VINS-Mono \cite{qin2018vins}. The default setting of each method is used for evaluation. For ORB-SLAM3, we select the monocular camera mode (without IMU) with different types of cameras: a pinhole camera, a fish-eye camera, and a thermal camera for evaluation (denoted by ORB3-Pinhole, ORB3-Fisheye, ORB3-Thermal respectively). 
	For LiDAR SLAM, we evaluate A-LOAM \cite{zhang2014loam}, LeGO-LOAM \cite{shan2018lego}, LINS \cite{qin2020lins} , and LIO-SAM \cite{shan2020lio}.

	The quantitative results are shown in Table \ref{ate rmse tab}. The estimated trajectories are visualized within the ENU frame as shown in Figure \ref{ate rmse fig}. As the robot travels on the ground, the visualization of most sequences is in 2D. 
	The results show that LiDAR-based methods outperform vision-based methods generally, especially in large-scale outdoor scenarios, but both kinds of methods do not perform well in certain cases. We discuss the results in detail as follows.
	
	\paragraph{Low illumination}  
	Sequence \emph{Roomdark06} and \emph{Street07} are in environments with low illuminations. ORB-SLAM3 using both pinhole (ORB3-Pinhole) and fisheye (ORB3-Fisheye) cameras fail in those sequences. Though ORB-SLAM3 adopts the strategy of adaptive histogram equalization to address bad illuminations, it failed to extract enough feature points in those dark scenes. Moreover, most extracted feature points were from far-away bright objects like street lamps or light screens, leading to significant estimation errors. By contrast, using a thermal-infrared camera, ORB-SLAM3 achieved significantly better robustness in the same scene because thermal cameras can distinguish objects under low visibility. However, we observe that some objects that can be well recognized by an RGB camera may appear texture-less in a thermal-infrared camera, for example, a flat colorful curtain. This phenomenon indicates that thermal-infrared cameras do not necessarily perform better than ordinary RGB cameras in some scenes.
	
\begin{figure}[h]
		\centering
		
		\includegraphics[width=0.44\linewidth]{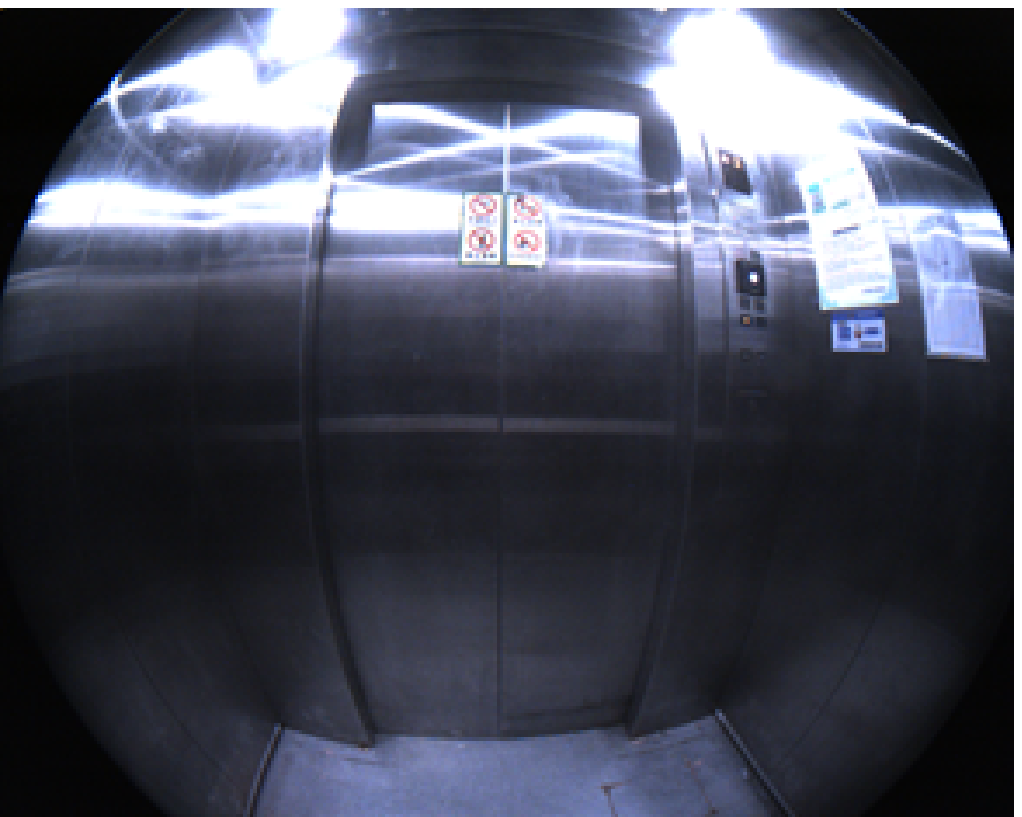}
		\includegraphics[width=0.44\linewidth]{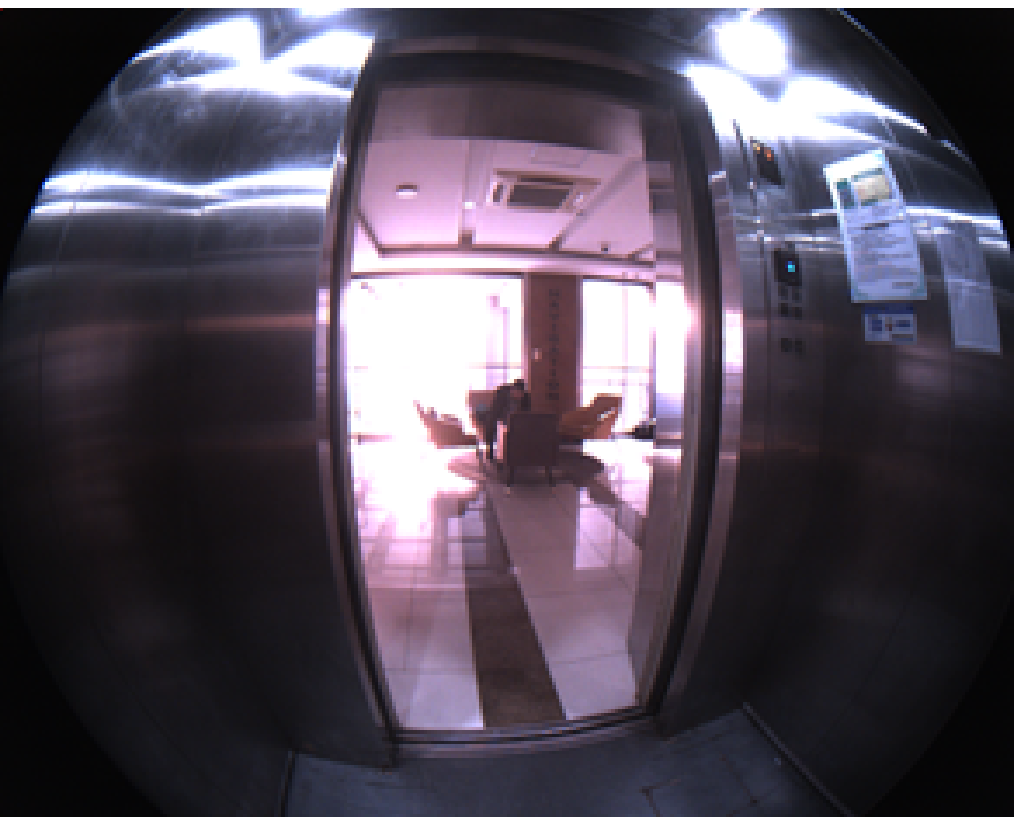}
		\caption{The images captured by a fish-eye camera when the robot  is leaving a lift.}
		\label{lift}
	\end{figure}

	\paragraph{Entering and leaving the lift}
	Sequence \emph{lift04} is a test sequence where the robot takes a lift to move across different floors as shown in Figure \ref{lift}. 
	Within a lift, SLAM systems with pure visual or laser information will consider the robot as being static, and only those with inertial information may recognize the robot moving upwards or downwards. Unfortunately, as shown in Figure \ref{ate rmse fig} (g) and (h), none of the tested SLAM systems succeeded in tracking the whole trajectory or reconstructing a complete map. Particularly, LIO-SAM drifts severely after the robot enters the lift due to a mismatch between IMU pre-integration and LiDAR odometry. As working on different floors is quite common for robots in daily life, it is meaningful and urgent to address the localization problem when robots take a lift to switch floors.
	
	\paragraph{Outdoor-Indoor switching}
	In Sequence \emph{Door01}, GNSS signals degraded drastically as the robot got closer to the door. All the SLAM systems can complete this sequence, while those with pure visual inputs yield large errors. We evaluate the GNSS SPP (single point positioning) performance by using a GNNS localization software RTKLIB \cite{takasu2009development}. Though the APE \cite{sturm2012benchmark} of RTKLIB still seems normal, actually indoor localization has failed, as Figure \ref{ate rmse fig} (f) shows. 
	GNSS positioning has a stable accuracy in environments with good satellite visibility, but they only work outdoors and usually suffer from signal loss, satellite ephemeris error, and multi-path effect. To better make use of GNSS signals, we provide a sky-pointing camera to apply the possible solutions such as \cite{marais2014toward}.

	\paragraph{Dynamic motion}
	Sequence \emph{Street07} was collected in a zigzag route with abrupt motions such as quick turning, braking, as well as speeding up and down. Neither visual SLAM nor LiDAR SLAM worked well on this test, as the results in Table \ref{ate rmse tab} show. Most Visual SLAM methods (ORB3-Pinhole, ORB3-Fisheye, CubemapSLAM) failed. LiDAR SLAM also produced large ATE errors in this case.
	
	\paragraph{Multiple cameras}
	Multiple-camera visual SLAM can take advantage of images in a wider field of view. We tested Multicol-SLAM \cite{urban2016multicol} with three fisheye cameras (two in the front, one on the left side), but it lost its track in almost every sequence. The reason might be that it tries to extract and match features directly from highly distorted images, which may easily cause false matches\cite{wang2018cubemapslam}. 

	The results indicate that the state-of-the-art  SLAM systems, both visual and LiDAR ones, may perform well on existing benchmark tests, they still require significant improvement to be applied to ground robots in daily life.  The results also indicate our dataset is a valid and valuable test field for existing SLAM systems. With rich sensory information and various scenarios, we believe our benchmark will promote the progress of robot navigation solutions.
	
	\section{Conclusion}
	
	We release M2DGR, a large-scale multi-sensor dataset focusing on ground robots' localization and mapping tasks. Our dataset contains a large pool of sensory information to encourage breakthroughs in multi-sensor fusion on SLAM. Furthermore, we tested and evaluated a few state-of-the-art SLAM systems based on our dataset and analyzed the defects and limitations of existing systems in different scenarios, which may point out potential developing directions for SLAM. In the future, we plan to update and extend our project from time to time, striving to build a comprehensive SLAM benchmark similar to the KITTI dataset \cite{geiger2013vision} for ground robots.


	\bibliographystyle{IEEEtran}
	\bibliography{ref}

\end{document}